%% file: main.tex
\documentclass[10pt,twocolumn,letterpaper]{article}

\usepackage{wacv}
\usepackage{times}
\usepackage{epsfig}
\usepackage{graphicx}
\usepackage{amsmath}
\usepackage{amssymb}
\usepackage{booktabs}
\usepackage{algorithm}
\usepackage{algorithmic}
\usepackage{enumitem}
\usepackage{tabu, multirow}
\usepackage[accsupp]{axessibility}

\wacvfinalcopy 

\ifwacvfinal
\usepackage[breaklinks=true,bookmarks=false]{hyperref}
\else
\usepackage[pagebackref=true,breaklinks=true,colorlinks,bookmarks=false]{hyperref}
\fi

\begin{document}

\title{GaIA: Graphical Information Gain based Attention Network for Weakly Supervised Point Cloud Semantic Segmentation}

\newcommand*\samethanks[1][\value{footnote}]{\footnotemark[#1]}

\author{Min Seok Lee\thanks{Equal contribution.} \;\;\;\;\; Seok Woo Yang\samethanks \;\;\;\;\; Sung Won Han\thanks{Corresponding author.}\\
School of Industrial and Management Engineering, Korea University\\
{\tt\small \{karel, joshy, and swhan\}@korea.ac.kr}}
\maketitle
\thispagestyle{empty}

\begin{abstract}
While point cloud semantic segmentation is a significant task in 3D scene understanding, this task demands a time-consuming process of fully annotating labels. To address this problem, recent studies adopt a weakly supervised learning approach under the sparse annotation. Different from the existing studies, this study aims to reduce the epistemic uncertainty measured by the entropy for a precise semantic segmentation. We propose the graphical information gain based attention network called GaIA, which alleviates the entropy of each point based on the reliable information. The graphical information gain discriminates the reliable point by employing relative entropy between target point and its neighborhoods. We further introduce anchor-based additive angular margin loss, ArcPoint. The ArcPoint optimizes the unlabeled points containing high entropy towards semantically similar classes of the labeled points on hypersphere space. Experimental results on S3DIS and ScanNet-v2 datasets demonstrate our framework outperforms the existing weakly supervised methods. We have released GaIA at \url{https://github.com/Karel911/GaIA}.
\end{abstract}

\section{Introduction}
Point cloud semantic segmentation is a fundamental task in the field of computer vision. With the success of deep neural networks, large-scale point cloud semantic segmentation on the 3D scene has drawn more attention due to its wide applications (e.g., augmented/virtual reality, autonomous driving, and robotics). However, a fully supervised method for point cloud semantic segmentation requires well-labeled point-wise annotations, and this entire process of data annotation is expensive \cite{qi2017pointnet,qi2017pointnet++,Tchapmi2017SEGCloudSS,Li2018PointCNNCO,choy20194d,wang2019dynamic,hu2020randla,thomas2019KPConv,hu2020randla,jiang2020pointgroup,zhao2021point,nekrasov2021mix3d,tang2022contrastive}.
\begin{figure}[ht]
\begin{center}
\includegraphics[scale=0.32]{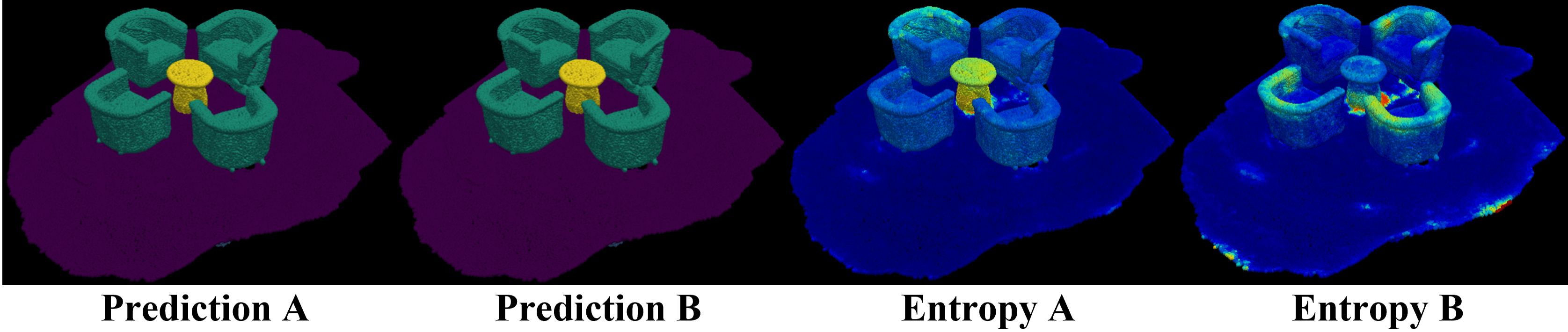}
\end{center}
\caption{Comparison of performance recognition and information uncertainty. Prediction of the network A has higher uncertainty in the table compared with the network B.}
\label{intro}
\end{figure}
To address this issue, recent studies have adopted a weakly supervised learning approach to train networks with partial annotations of point clouds. Previous studies \cite{wei2020multi,cheng2021sspc,xu2020weakly,zhang2021weakly,zhang2021perturbed,hou2021exploring,liu2021one,Yang_2022_CVPR,Li_2022_CVPR} improved the semantic segmentation performance close to that of fully supervised one on small-scale datasets (e.g., ShapeNet \cite{chang2015shapenet} and PartNet \cite{mo2019partnet}) as well as large-scale datasets (e.g., S3DIS \cite{armeni20163d} and ScanNet-v2 \cite{dai2017scannet}).

In contrast to existing studies, this study focuses on alleviating epistemic uncertainty to obtain high-quality feature representations under sparse annotation. In Fig. \ref{intro}, if two networks show a similar performance or visualization result, it is hard to determine which network is semantically well-embedded. To observe whether there is a difference in the estimation of the two networks, Shannon entropy \cite{shannon1948mathematical} was employed for epistemic uncertainty quantification \cite{kendall2017uncertainties,malinin2018predictive}. In measuring the entropy of each point, it was observed that the reliability of the network prediction may differ even if the same result is obtained. Starting from this experimental result, the question was raised as to whether alleviating epistemic uncertainty improves segmentation performance along with satisfactory point cloud embedding. To address epistemic uncertainty reduction, we introduce two approaches: reducing the entropy of each point and effective optimization for points containing high entropy.

Reducing epistemic uncertainty is regarded as alleviating the entropy of each sample \cite{harpur1996development,harpur1997low,smith2018understanding}. To reduce the entropy of each point, we treat points with low entropy as credible information to update the probability distribution of points containing high entropy. Reliable points near the ambiguous decision boundary of the network are identified by measuring relative entropy, because not all reliable points are important. As a relative entropy measure, this study introduces graphical information gain, which is determined by the relative entropy between the entropy of the target point and that of its neighborhood. When a point has entropy lower than that of its neighborhood, it is more reliable. Based on the reliability, we enhance the point representation and update the point including the high entropy by propagating the credible information to the uncertain points.

Under the sparse annotation, effective optimization of the unlabeled points is important for achieving satisfactory semantic segmentation. Existing studies organize the relation network \cite{xu2020weakly,zhang2021perturbed} or class prototypical matrix \cite{zhang2021weakly} to optimize the unannotated points. For loss computation, the softmax function is widely employed to present class probability. However, the softmax has a limitation in terms of data optimization in that it can neither explicitly enhance the similarity of intra-class features nor discriminate inter-class features \cite{deng2019arcface}. Moreover, previous studies equally focused on all unlabeled points during the optimization process. Although the points containing low entropy are semantically well-embedded in the optimization process, the network should focus more on optimizing the unannotated points with high entropy to improve segmentation performance. Therefore, it is necessary to overcome the drawbacks of softmax and address the optmization of highly uncertain points.

This study proposes a graphical information gain-based attention network (GaIA) for weakly supervised point cloud semantic segmentation. GaIA aims to reduce epistemic uncertainty using the graphical information gain and the anchor-based additive angular margin loss called ArcPoint. The graphical information gain measures the relative entropy between the entropy of the target point and that of its neighborhoods to discriminate reliable information. Based on relative entropy, GaIA updates the feature embedding of the unlabeled points containing high entropy toward semantically similar embedding of the labeled points. To address the limitation of softmax and focus on unlabeled point optimization, we introduce ArcPoint loss. By penalizing the unannotated points with high entropy using an additive angular margin in loss computation, ArcPoint optimizes the uncertain points embedded in the hypersphere toward a semantically similar embedding of the labeled points. The main contributions of this study are as follows:
\begin{itemize}[leftmargin=\dimexpr\parindent+0.1mm+0.1\labelwidth\relax]
    \item Epistemic uncertainty reduction is studied to improve weakly supervised point cloud semantic segmentation performance. To the best of our knowledge, this is the first approach to focus on epistemic uncertainty reduction for a performance gain in the weakly supervised point cloud semantic segmentation.
    
    \item For epistemic uncertainty reduction, we propose the graphical information gain to measure the relative entropy between the entropy of target point and that of its neighborhoods to identify reliable information.
    
    \item The proposed ArcPoint loss contributes to the epistemic uncertainty reduction by enabling the network to embed the unlabeled points with high entropy toward the reliable labeled points.

    \item GaIA improves mIoU by 2.2\%p and 4.4\%p compared with existing weakly supervised learning methods on two benchmark datasets (e.g., S3DIS and ScanNet-v2) under the 1 and 20 pts annotation.
 \end{itemize}


\begin{figure*}[ht]
\begin{center}
\includegraphics[scale=0.34]{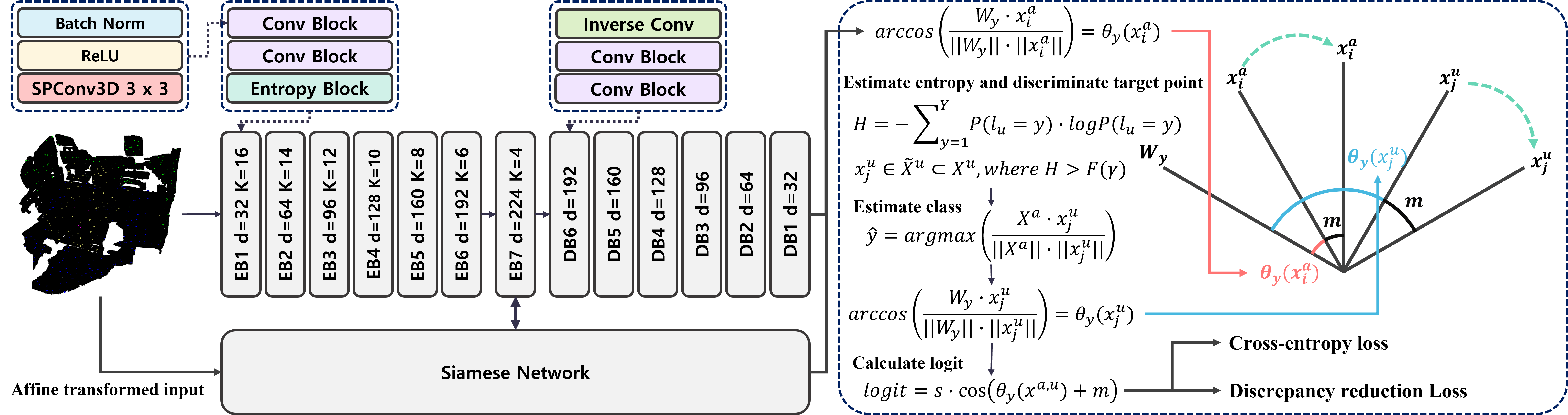}
\end{center}
  \caption{Overall architecture.}
\label{arch}
\end{figure*}

\section{Related work}
\subsection{Weakly supervised semantic segmentation on point cloud}
Studies on 3D point cloud semantic segmentation have improved performance using fully annotated supervision learning \cite{qi2017pointnet,qi2017pointnet++,Tchapmi2017SEGCloudSS,Li2018PointCNNCO,choy20194d,wang2019dynamic,hu2020randla,thomas2019KPConv,hu2020randla,jiang2020pointgroup,zhao2021point,nekrasov2021mix3d,tang2022contrastive}. Despite this achievement, annotating all point clouds remains a time-consuming task. To address this problem, recent studies have adopted a weakly supervised learning approach. Weakly supervised point cloud semantic segmentation performs segmentation with partial annotations for the point cloud. Existing studies generated semantically transformed types of point clouds, such as 2D segmentation maps \cite{wang2020weakly}, subcloud-level annotation \cite{wei2020multi}, and superpoint \cite{cheng2021sspc}. With sparse annotation, previous approaches have employed pre-training method \cite{hou2021exploring, zhang2021weakly}, contrastive learning \cite{hou2021exploring, liu2021one, Li_2022_CVPR}, and learning distribution consistency \cite{xu2020weakly, zhang2021perturbed, Li_2022_CVPR, Yang_2022_CVPR} to learn spatial information of point clouds. For learning the topology of a point cloud, graph-structure was utilized to represent features of points \cite{cheng2021sspc, liu2021one, zhang2021perturbed}. Different from the previous approaches, we propose a novel weakly supervised framework that aims to reduce network uncertainty and effectively optimize unlabeled points.

\subsection{Uncertainty quantification and reduction}
Uncertainty quantification is important for precise decision-making in various domains \cite{amodei2016concrete}, such as autonomous driving \cite{feng2018towards,choi2019gaussian} or medical image analysis \cite{labonte2019we,seebock2019exploiting,roy2019bayesian,nair2020exploring,reinhold2020validating}. Uncertainty in the predictive process is caused by three components: data uncertainty, epistemic uncertainty, and distributional uncertainty \cite{quinonero2008dataset,gal2016uncertainty,kendall2017uncertainties,malinin2018predictive}. Among these three types of uncertainty, this study focuses on epistemic uncertainty, which measures the information uncertainty in predicting network parameters given the data \cite{gal2016uncertainty,malinin2018predictive}. Uncertainty can be reduced, such that the lower the uncertainty, the higher is the network performance is \cite{harpur1996development,harpur1997low,malinin2018predictive}. Based on this property, we introduce a network that focuses on epistemic uncertainty reduction to improve point cloud semantic segmentation performance. For the uncertainty quantification measure, Shannon entropy, which represents information uncertainty \cite{shannon1948mathematical}, is adopted, whereby the entropy of each point is estimated to identify reliable information. Our approach updates uncertain points near the ambiguous decision boundary of the network by propagating credible features.

\subsection{Sparse annotation embedding}
Using point cloud data is a more attractive approach compared to a transformed representation (e.g., a voxel or mesh). However, it is difficult to employ raw point clouds due to their disordered and unstructured properties \cite{qi2017pointnet,engelmann2017exploring}. Furthermore, it is challenging to generate a high-quality feature representations from partially annotated point clouds. Thus, existing studies focus on the feature representation of labeled points shared with unlabeled point clouds \cite{xu2020weakly,zhang2021weakly,zhang2021perturbed, liu2021one, Li_2022_CVPR, Yang_2022_CVPR}. To obtain the feature embedding, previous studies minimize the difference between the ground truth and projected label \cite{wang2020weakly,zhang2021weakly}. In addition to the aforementioned studies, other approaches optimize the divergence between two probabilistic distributions \cite{xu2020weakly,zhang2021perturbed, liu2021one, Li_2022_CVPR, Yang_2022_CVPR}. In the training process, the above studies employed the softmax function. However, the softmax has a limitation when classifying an open-data set in that it is not in the training data \cite{deng2019arcface}. Thus, the convergence of intra-class data and divergence of inter-class data should be enhanced to effectively embed unfamiliar data. Moreover, previous studies have optimized all unlabeled data equally. In contrast to these studies, we focus on the optimization of unlabeled points with high entropy for effective optimization. The uncertain points are closely embedded along with the semantically similar labeled points on the hypersphere by using the labeled points as anchor.

\section{Method}
\subsection{GaIA overview}
{\bf Architecture: }GaIA is designed to alleviate epistemic uncertainty. To reduce the high entropy of the uncertain points, we organize the entropy block and ArcPoint loss. As depicted in Fig. \ref{arch}, 3D U-Net is implemented as a backbone network with sub-manifold sparse convolution and sparse convolution as in \cite{graham20183d,jiang2020pointgroup}. Input $X$ is a point set of $N$ points. Each point $x_i \in \mathbb{R}^{6}$ is represented by a concatenation of 3D coordinates and RGB colors, where $i\in \{1, ... ,N\}$. Then, $X$ is voxelized to a size of 0.02m. The semantic features are extracted by feeding $X$ to a couple of convolution and entropy blocks. Each convolution block comprises a sequence of batch normalization-ReLU-sparse convolutional operations (SPConv3D). Subsequently, the entropy block computes the entropy variation between the target point entropy and entropy of its neighborhoods, referred to as graphical information gain in this study. As an attention weight, graphical information gain enhances reliable point representation and propagates the information to their neighborhoods. After extracting semantic features from the encoder blocks, $X$ is reconstructed using a decoder. The entropy block at the decoder is excluded because applying the entropy block to each decoder block results in computational inefficiency. In fact, when a decoder with an entropy block is organized, the inefficiency increases with respect to the performance gain.

{\bf Learning strategy: }To embed the unlabeled points, we adopt a Siamese network branch \cite{bromley1993signature,koch2015siamese} to GaIA. The Siamese branch maintains the consistency between the prediction of the original input $X$ and that of the affine transformed input $aff(X)$. This learning strategy improves the embedding performance by imposing constraints on unannotated points \cite{xu2020weakly}. For the affine transformation of a given input point $X$, we apply a random noise, flipped with the $x$ and/or $y$ axes, and rotated at random angles to the $x$ axis. Subsequently, to achieve a more robust network against the sparse annotation, we impose more constraints on $X$ by employing an elastic distortion. Initially, GaIA is trained excluding the Siamese branch for 100 epochs because the constraints result in unstable entropy for each point at earlier stages. After optimizing the original network, we adopt the Siamese branch to minimize the discrepancy between network predictions.

\begin{algorithm}[ht]
\caption{Entropy block operation}
\label{graphical_IG}
\begin{algorithmic} [1]
\STATE {\bfseries Input: }Point cloud representation $X \in \mathbb{R}^{N \times d}$.
\STATE Initialize: $\widetilde{X} = \mathcal{F}(X), \; where \; \widetilde{X} \in \mathbb{R}^{N \times Y}$ and \\ graph $G(N, E)$ $\leftarrow KNN(X_{loc}, \; K)$.

\STATE Get $H$: $H_i = -\sum_{y=1}^{Y}P(x_i=y) \cdot logP(x_i=y),$ \\ 
$where \; P(x_i) = softmax(x_i)$ and $x_i \in \widetilde{X}$.


\STATE Calibrate: $\widetilde{H}_i = \sum_{j\neq{i}}^{k}{(D_{i,j})^{-2}} \cdot H_{j} / \sum_{j\neq{i}}^{k}{(D_{i,j})^{-2}},$ \\
$where \; x_j \in neighbor(x_i)$.

\STATE Get $GI$: $GI_i = |H_i - \widetilde{H}_i|$.

\STATE Neighbor aggregation: $x_i^{n} = (\sum_{j\neq{i}}^{k}{x_{j} \otimes GI_{j}}) / K$.

\STATE Update point embedding: $\widetilde{X} = \widetilde{X} + (\widetilde{X} \otimes GI) + \widetilde{X}^{N},$ \\
$where \; x_i^{n} \in \widetilde{X}^{N}$.

\STATE {\bfseries Output: }$O = \mathcal{F}(\widetilde{X}), \; where \; O \in \mathbb{R}^{N \times d}$.

\end{algorithmic}
\end{algorithm}

{\bf Optimization: }Existing studies \cite{xu2020weakly,zhang2021weakly,zhang2021perturbed, liu2021one, Li_2022_CVPR, Yang_2022_CVPR} employed softmax cross-entropy loss and treat the points equally for network optimization. Different from the previous approaches, this study focuses on optimizing the points including high entropy. Inspired by ArcFace \cite{deng2019arcface}, which addresses the limitation of conventional softmax cross-entropy loss, is adopted as a baseline form of the loss function. However, ArcFace loss cannot deal with a large number of unannotated points because it requires the ground truth in the training phase. Thus, we propose an anchor-based additive angular margin loss called ArcPoint. ArcPoint loss aims to embed the unlabeled points toward semantically similar points by employing labeled points regarded as anchors. In Fig. \ref{arch}, ArcPoint at first optimizes the distance $\theta_y(x_i^a)$ between the class-prototypical weight matrix $W_y$ and annotated point $x_i^a \in X^a$ on the normalized hypersphere. Afterward, unannotated points containing high entropy $x_j^u \in \widetilde{X}^u$ are identified and the angle between $X^a$ and $x_j^u$ is computed. Here, $X^a$ functions as an anchor in leading the $x_j^u$ towards the nearest class $W_y$. A more detailed computing process for ArcPoint loss is demonstrated in Section \ref{loss}.

\subsection{Graphical information gain}
Graphical information gain ($GI$) measures the relative entropy between the entropy of the target point and that of its neighbors to identify reliable information. $GI$ is theoretically based on information uncertainty \cite{shannon1948mathematical}.
\begin{figure}[ht]
\begin{center}
\includegraphics[scale=0.29]{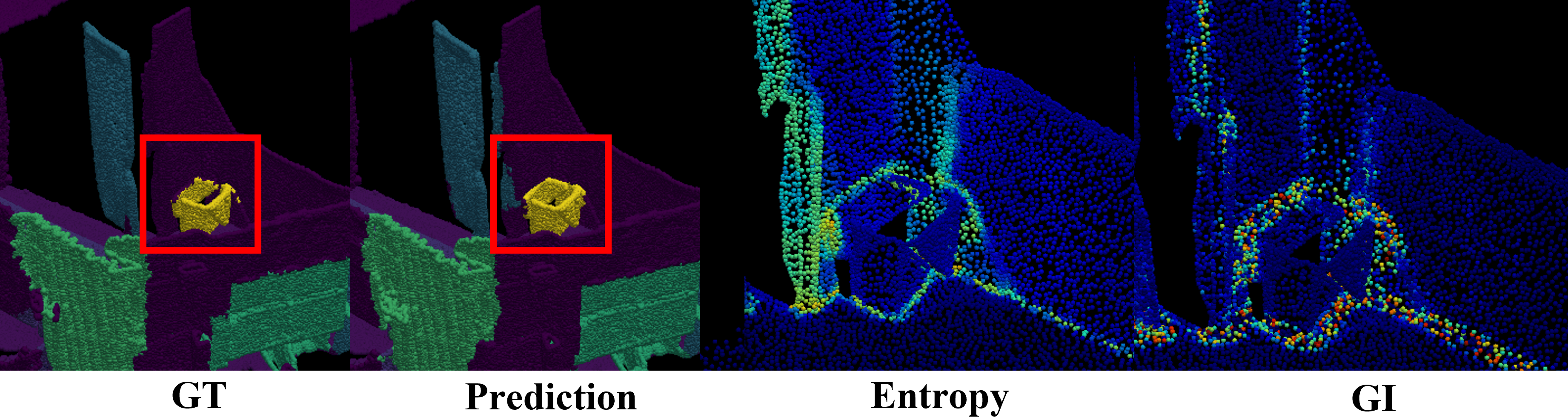}
\end{center}
\caption{Visualization of a decision boundary and graphical information gain. Red points indicate high entropy and GI values.}
\label{decision_boundary}
\end{figure}
\begin{figure}[ht]
\begin{center}
\includegraphics[scale=0.275]{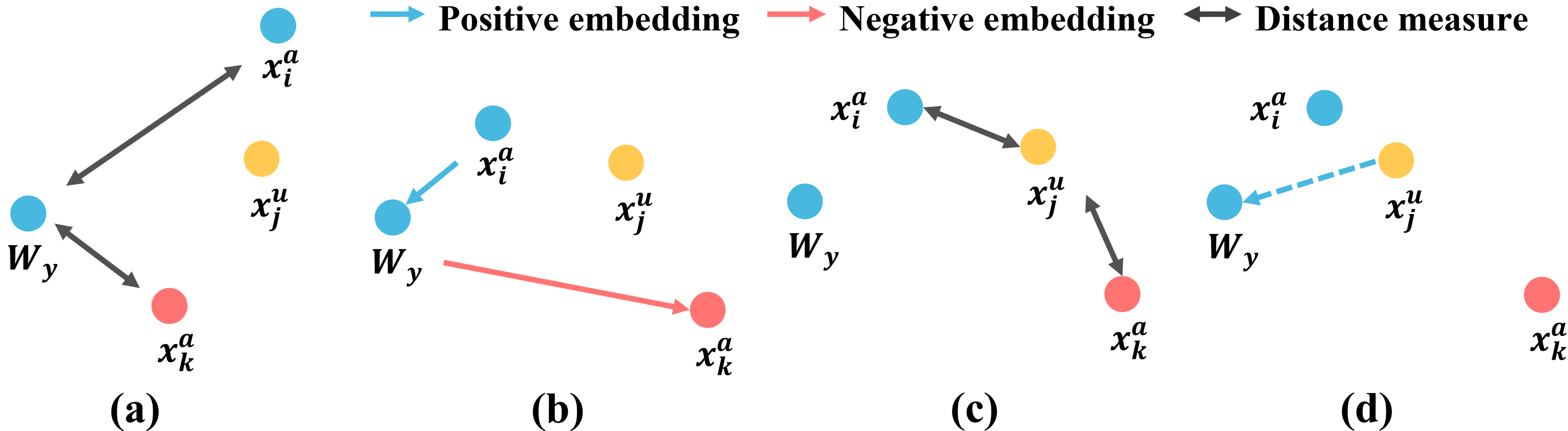}
\end{center}
\caption{Embedding process of the ArcPoint loss.}
\label{embedding_process}
\end{figure}
The entropy $H$ represents the information uncertainty using the probability of event $i$ as follows: $H = -\sum_i{P_i \cdot logP_i}$. That is, if the probability distribution of the classes is sparse, a network can make a reliable decision for class prediction. Focusing on this property, entropy is utilized through three phases to alleviate epistemic uncertainty: \lowercase\expandafter{\romannumeral1}) measure entropy of each point, \lowercase\expandafter{\romannumeral2}) compute graphical information gain, and \lowercase\expandafter{\romannumeral3}) update the point embeddings with reliable representations. As shown in Algorithm \ref{graphical_IG}, the input point cloud $X \in \mathbb{R}^{N \times d}$ is projected onto $\widetilde{X} \in \mathbb{R}^{N \times Y}$ using the SPConv3D operation $\mathcal{F}(\cdot)$, where $Y$ denotes the number of classes. In addition, based on the coordinates $X_{loc}$, the k-nearest neighbor algorithm is applied to the input $X$ to identify the neighborhood. In line 3, entropy $H_i$ for each point $x_i$ is computed. To obtain $GI$, we aggregate the entropy of neighborhoods $H_j$ that are inversely proportional to the Euclidean distance ($D_{i,j}$) between target point $x_i$ and its neighborhood $x_j$ in line 4. Inverse $D_{i,j}$ imposes more weights on the neighborhood entropy, which is geometrically close to the target $x_i$. In line 5, $GI$, which is regarded as relative entropy, is obtained by subtracting the calibrated entropy $\widetilde{H}$ from the original entropy $H$. When the target point contains lower entropy compared to that of its neighborhoods, the results are more reliable. As depicted in Fig. \ref{decision_boundary}, it is recognized that the $GI$ highlights reliable points with low entropy near the ambiguous decision boundary of the network. Subsequently, we enhance the reliable representations using $x_i \otimes GI_i$, and neighborhood information is aggregated along with normalization in lines 6 and 7. Based on both enhanced representations, the point embeddings are updated to reduce epistemic uncertainty. Finally, the entropy block reconstructs the updated representation $\widetilde{X} \in \mathbb{R}^{N \times C}$ to $O \in \mathbb{R}^{N \times d}$ by using the sparse convolutional operation $\mathcal{F}(\cdot)$. Further analysis of the $GI$ is offered in the Supplementary-analysis on graphical information gain.

\begin{algorithm}[ht]
\caption{Anchor based additive angular margin loss}
\label{algoritm_arcpoint}
\begin{algorithmic} [1]
\STATE {\bfseries Input: }labeled anchor $x_i^a \in X^a$, unannotated points $X^u \in \mathbb{R}^{N \times d}$, $y^{th}$ class-prototypical weight matrix $W_y \in \mathbb{R}^{d}$, re-scaler $s$, and margin parameter $m$.

\STATE Get angle and add angular margin:\\
$\theta_{y}(x_i^a) + m = arccos(\frac{W_y^\intercal \cdot x_i^a}{||W_y|| \cdot ||x_i^a||}) + m$

\STATE Calculate $H$: $H = -\sum_{y=1}^{Y}P(l_u=y) \cdot logP(l_u=y),$\\
$where \;\; P(l_u) = softmax(s \cdot (\frac{X^u \cdot W}{||X^u|| \cdot ||W||}))$, and $W \in \mathbb{R}^{d \times Y}$

\STATE Discriminate the points containing high entropy:\\
$x_j^u \in \widetilde{X}^u \subset X^u, \; where \; H > F(\gamma)$

\STATE Estimate the nearest anchor: $\hat{y} = argmax(\frac{X^a \cdot x_j^u}{||X^a|| \cdot ||x_j^u||})$

\STATE Add angular margin for unannotated points:\\
$\theta_{y}(x_j^u) + m = arccos(\frac{W_{\hat{y}}^\intercal \cdot x_j^u}{||W_{\hat{y}}|| \cdot ||x_j^u||}) + m$

\STATE Calculate final logit: \\
$l =
\begin{cases}
s \cdot cos(\theta_{y}(x^i) + m), & \mbox{if\;\;} i \in \{a,u\}, \; where\\
& \mbox{$x^{a} \in X^{a} \;\; and \;\; x^{u} \in \widetilde{X}^{u}$}\\
s \cdot cos(\theta_{y}(x^u)), & \mbox{otherwise}
\end{cases}$

\STATE {\bfseries Output: }Final logit $l$
\end{algorithmic}
\end{algorithm}

\subsection{Loss function design}\label{loss}
{\bf Anchor based additive angular margin loss: }ArcPoint is designed to effectively embed unannotated points by addressing the limitation of conventional softmax cross-entropy and ArcFace losses. Fig. \ref{embedding_process}-(a) and -(b) illustrate the original embedding process of ArcFace which embeds the intra-class anchor point similarly while discriminating the inter-class point. Following both (a) and (b), the angles between the unlabeled points and anchors are measured, including different classes. Subsequently, in Fig. \ref{embedding_process}-(d), the nearest anchor of each unannotated point is determined by the smallest angle. Afterward, the unannotated point embedding is relocated toward the class which is the same as the nearest anchor. The detailed embedding process is presented in Algorithm \ref{algoritm_arcpoint}.

In line 2, the $i^{th}$ labeled anchor point $x_i^a \in \mathbb{R}^{d}$ belonging to the class $y$ is embedded on the hypersphere by computing the angle $\theta_{y}(x_i^a)$ between $x_i^a$ and $W_y$. Here, $W_y$ denotes the $y^{th}$ column of class-prototypical weight matrix $W \in \mathbb{R}^{d \times Y}$. The angle with an additive angular margin $m$ is penalized to enhance intra-class intensity and inter-class distinction. To optimize the unannotated points along with epistemic uncertainty reduction, we focus on the points with high entropy. In lines 3 and 4, the entropy of unannotated points is calculated using a re-scaled logit $l_u$ to discriminate the target points $\widetilde{X}^u$ that contain high entropy. Here, the function $F(\gamma)$ denotes the $\gamma$ quantile of $H$, such that the higher area of $\gamma$ in the distribution of $H$ is adopted. Subsequently, to estimate the nearest anchor regarded as a class, we measure $cos\theta$ between the entire anchor $X^a$ and target point $x_j^u$, which is the $j^{th}$ instance of the target points $\widetilde{X}^u$. Following the estimation, the angular margin $m$ is added to the angle $\theta_{\hat{y}}(x_j^u)$ measured by the estimated class weight $W_{\hat{y}}$ and $x_j^u$ in line 6. For the final logit calculation, both $\theta_{y}(X^a)$ and $\theta_{y}(\widetilde{X}^u)$ are applied to the margin, except for the other cases. The logit passes the cross-entropy loss through a softmax function. In the inference phase, logit is computed without the additive angular margin as follows: $logit = s \cdot (\frac{X \cdot W}{||X|| \cdot ||W||})$. This optimization effect is discussed in Section \ref{ablation_arcpoint} by visualizing similarities between $X^a$ and $\widetilde{X}^u$ corresponding to each class.

{\bf Loss configuration: }The loss function $\mathcal{L}$ comprises the ArcPoint based both softmax cross-entropy loss $\mathcal{L}_{ce}$ and distribution discrepancy reduction loss $\mathcal{L}_{sia}$, as follows: $\mathcal{L} = \mathcal{L}_{ce} + \mathcal{L}_{ce}^{aff} + \mathcal{L}_{sia}$. In Eq (\ref{L_cls}), $A$ denotes the number of labeled points, and the annotated points are optimized by using the penalty term $m$. In Eq (\ref{siamese_loss}), when the Siamese branch is applied to GaIA, the segmentation loss $\mathcal{L}_{ce}^{aff}$ for the affine transformed input $aff(X)$ and distribution discrepancy reduction loss $\mathcal{L}_{sia}$, which are based on ArcPoint, are organized. The $\mathcal{L}_{sia}$ minimizes the L2 distance between all probabilistic predictions of the original network and those of the Siamese branch. In this process, the unannotated points are optimized. Here, the unlabeled points containing low entropy are not subject to the angular margin, but are involved in distance minimization.
\begin{equation}
\begin{gathered}
\label{L_cls}
\resizebox{1.\hsize}{!}{$
\mathcal{L}_{ce} = -\frac{1}{A}\sum\limits_{i=1}^{A}{log
\frac{e^{s \cdot cos(\theta_{y}(x_i^a) + m)}}{e^{s \cdot cos(\theta_{y}(x_i^a) + m)} + \sum_{j=1, j\neq y)}^{Y}{e^{s \cdot cos\theta_j}}}}
$}
\end{gathered}
\end{equation}

\begin{equation}
\begin{gathered}
\label{siamese_loss}
\resizebox{0.78\hsize}{!}{$
\mathcal{L}_{sia} = ||P(X) - P(aff(X))||_2, \;\;\; where$}\\
\resizebox{0.95\hsize}{!}{$
P(X) = \frac{1}{N}\sum\limits_{i=1}^{N}{\frac{e^{s \cdot cos(\theta_{y}(x_i^{a,u}) + m)}}{e^{s \cdot cos(\theta_{y}(x_i^{a,u}) + m)} + \sum_{j=1, j\neq y)}^{Y}{e^{s \cdot cos\theta_j}}}}
$}
\end{gathered}
\end{equation}


\section{Experiment}
\subsection{Experimental setup}\label{exp_setup}
{\bf Datasets: }S3DIS \cite{armeni20163d} contains 271 scenes for six areas from three different buildings consisting of 3D RGB point clouds. Each point was annotated with one of 13 semantic categories. All the classes were used in the instance evaluation. GaIA was evaluated on two settings: \lowercase\expandafter{\romannumeral1}) Area 5 is used for testing and all others are utilized for training, \lowercase\expandafter{\romannumeral2}) in the 6-fold cross validation each area is treated as the test set once. Experiments were also conducted on the ScanNet-v2 \cite{dai2017scannet} which consisted of 1,613 scenes annotated with 20 classes. The dataset was split into 1,201 training, 312 validation, and 100 test scenes. To make it comparable to other approaches,the benchmark results are reported for the official test set.

\input{table/benchmark_s3dis}
\begin{figure}[ht]
\begin{center}
\includegraphics[scale=0.33]{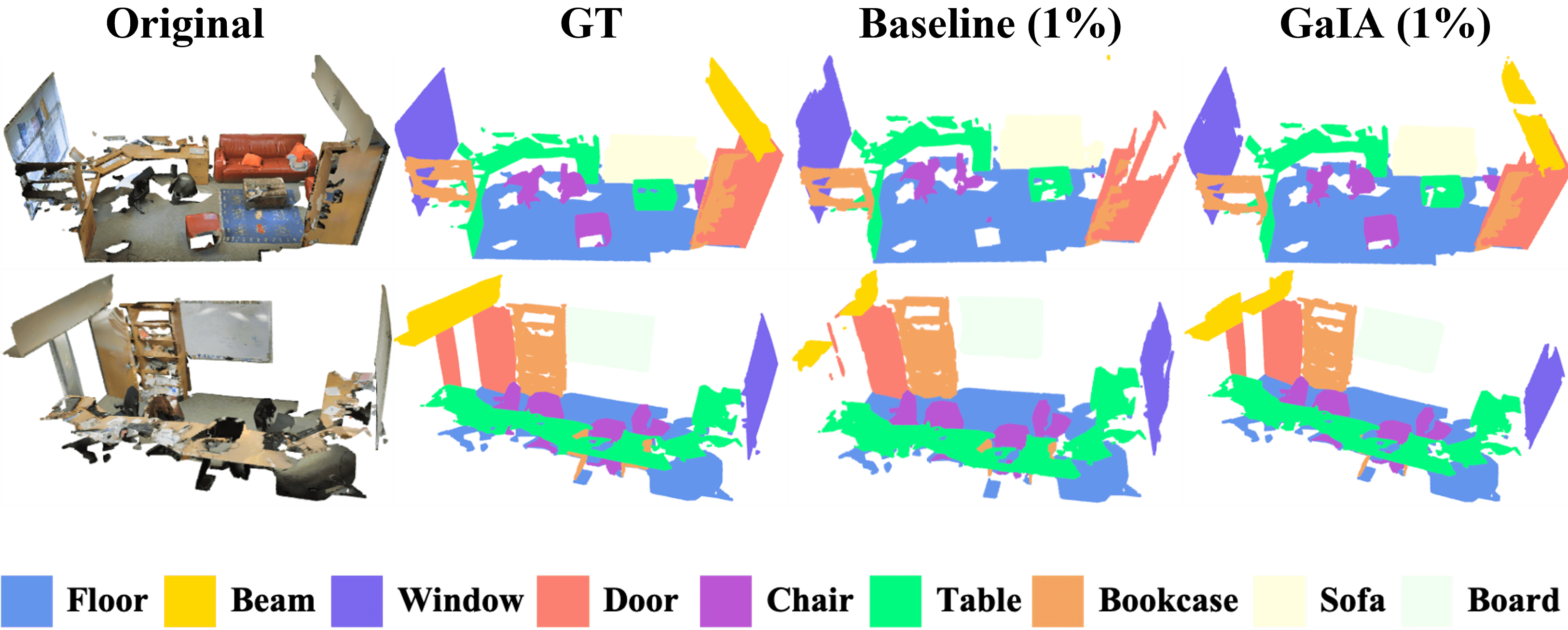}
\end{center}
\caption{Comparison of qualitative results on S3DIS.}
\label{visual_comparison_s3dis}
\end{figure}
  
{\bf Implementation details: }For a sparse annotation setting, the points corresponding to the supervision ratio (1pt, 20 pts, and 1\%) per class were labeled for each scene. GaIA was trained on a RTX A6000 GPU using the Adam optimizer with an initial learning rate of 0.01 and weight decay of 0.0001. The number of neighbors $K$ was initially set to 16 and then reduced by 4 following the encoder block. The angular margin $m$ was empirically determined to be 0.1 and re-scale factor $s$ was set to 16. For S3DIS, we set the batch size to 150, and a batch size of 8 was employed for ScanNet-v2. GaIA was implemented using the PyTorch framework. As evaluation metric, the mean intersection over union (mIoU) was adopted.

\input{table/benchmark_scannet}

\subsection{Experimental results}
{\bf S3DIS: }GaIA was compared with existing fully supervised (100\%) \cite{qi2017pointnet, qi2017pointnet++,Li2018PointCNNCO,thomas2019KPConv,hu2020randla,yan2020pointasnl,choy20194d,zhao2021point,hou2021exploring, tang2022contrastive} and weakly supervised (1pt and 1\%) \cite{xu2020weakly,zhang2021weakly,zhang2021perturbed, liu2021one, Yang_2022_CVPR, Li_2022_CVPR} methods on S3DIS area 5 and 6-Fold, as listed in Tab. \ref{benchmark_s3dis}. Under the 1pt and 1\% annotation on area 5, GaIA improved mIoU by 2.2\%p and 1.2\%p, respectively, compared to HybridCR \cite{Li_2022_CVPR}. In comparison of 6-Fold result on S3DIS, GaIA achieved the close performance on as that of the fully supervised state-of-the-art method \cite{zhao2021point} (-2.7\%p) and surpassed the existing weakly supervised method \cite{Li_2022_CVPR} (+1.6\%p). In Fig. \ref{visual_comparison_s3dis}, we visualized the qualitative results on S3DIS dataset. Compared to the baseline network excluded the Siamese branch, entropy blocks, and ArcPoint loss, GaIA precisely detected the classes, in particular, the beam and door. Additional visual comparison is reported in Supplementary.

\begin{figure*}[ht]
\begin{center}
\includegraphics[scale=0.6]{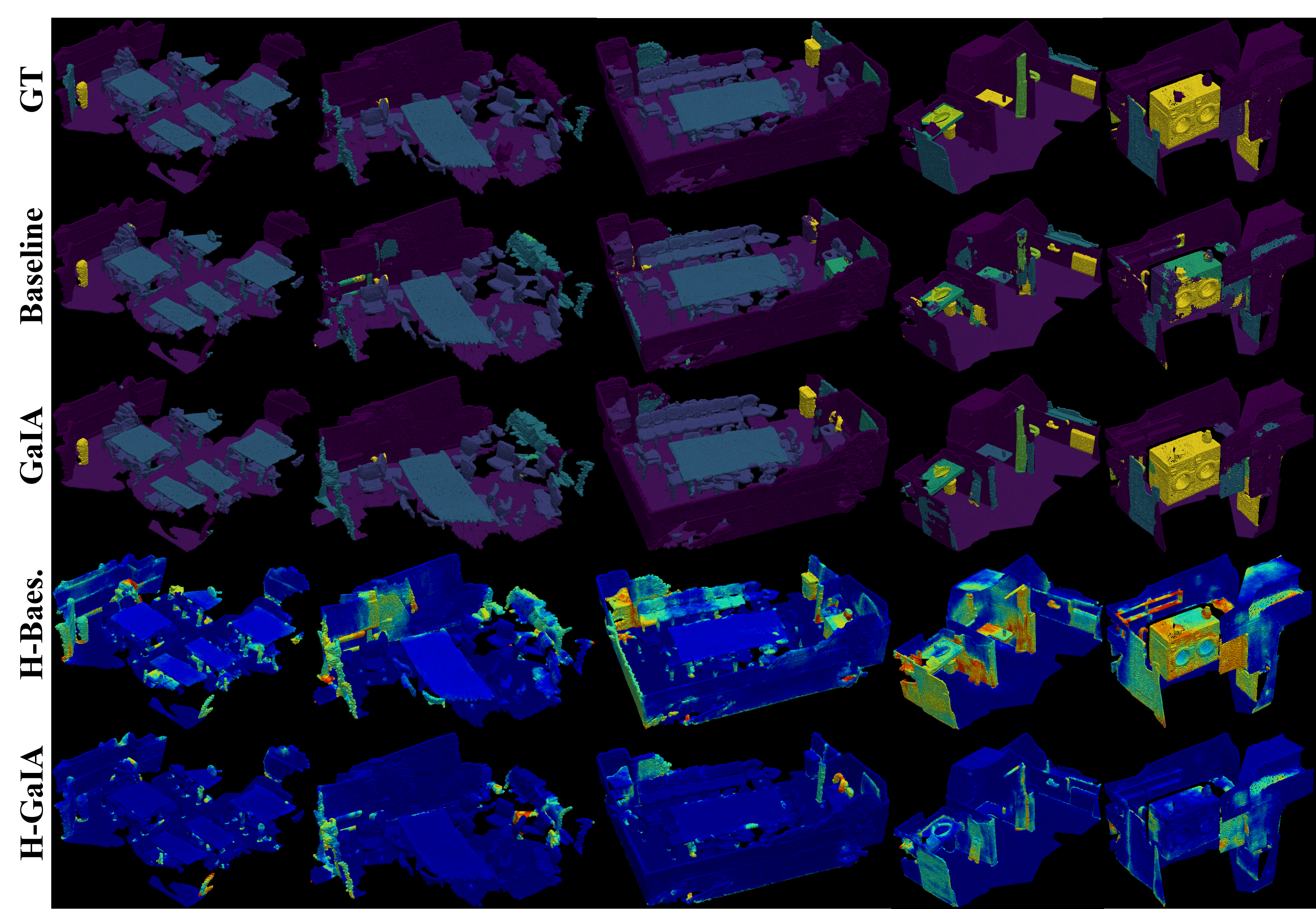}
\end{center}
\caption{Comparison of qualitative results on ScanNet-v2 validation set. $H$ denotes the entropy visualization.}
\label{visual_comparison}
\end{figure*}

{\bf ScanNet-v2: }The benchmark results of the ScanNet-v2 are listed in Tab. \ref{benchmark_scannet}. Compared to the existing weakly supervised methods, HybridCR \cite{Li_2022_CVPR}, GaIA improved mIoU by 8.4\%p under the 1\% annotation setting. Moreover, in limited annotations (LA) benchmark, GaIA outperformed Hou \etal \cite{hou2021exploring} (+8.3\%p), OTOC \cite{liu2021one} (+4.4\%p), and MIL Transformer \cite{Yang_2022_CVPR} (+9.4\%p). Remarkably, despite having more than 100$\times$ fewer annotations (1pt, 0.005\%), GaIA exhibited a performance surpassing that of Zhang \etal (+1.0\%p) and close to that of PSD (-2.6\%p). As depicted in Fig. \ref{visual_comparison}, GaIA was effective in epistemic uncertainty reduction compared to the baseline. When both networks exhibited similar segmentation results (cols 1 to 3), GaIA estimated the point cloud with higher reliability (rows 4 and 5). Although the segmentation results of both networks were unsatisfactory (cols 4 and 5), GaIA framework effectively alleviated epistemic uncertainty compared to the baseline.

\input{table/ablation_module}

\section{Ablation study}
\subsection{Effectiveness of the proposed components}
Ablation studies were conducted to analyze the contribution of each proposed component to performance gain. As listed in lines 1 and 2 of Tab. \ref{table_ablation_module}, the Siamese branch highly contributed to the performance gain compared with the baseline. This is because the Siamese branch is directly involved in the optimization of unlabeled points, which occupy the largest proportion of the data. This tendency was also observed in a previous study \cite{xu2020weakly}. When the proposed components were applied to the baseline with the Siamese branch, it was confirmed that the performance gain mainly originated from the entropy block (EB) compared with the ArcPoint loss (AP), as listed in lines 3 and 4. Under the 1pt annotation setting (0.005\%), employing both components improved the performance by 8.7\%p and 4.5\%p compared to the baseline and Siamese network, respectively. We demonstrated more detailed analysis on the proposed components in Supplementary.

\subsection{Effectiveness of ArcPoint loss}\label{ablation_arcpoint}
We conducted quantitative and qualitative experiments to validate the effectiveness of ArcPoint loss. In Tab. \ref{table_ablation_module}, ArcPoint loss was compared with the conventional softmax cross-entropy loss (line 4) and ArcFace (AF) loss (line 6). For a fair comparison, the Siamese branch with entropy blocks was employed. ArcFace outperformed the conventional softmax cross-entropy and L2 losses on 1pt and 1\% annotations with 0.3\%p and 0.4\%p gains, respectively.
\begin{figure}[ht]
\begin{center}
\includegraphics[scale=0.45]{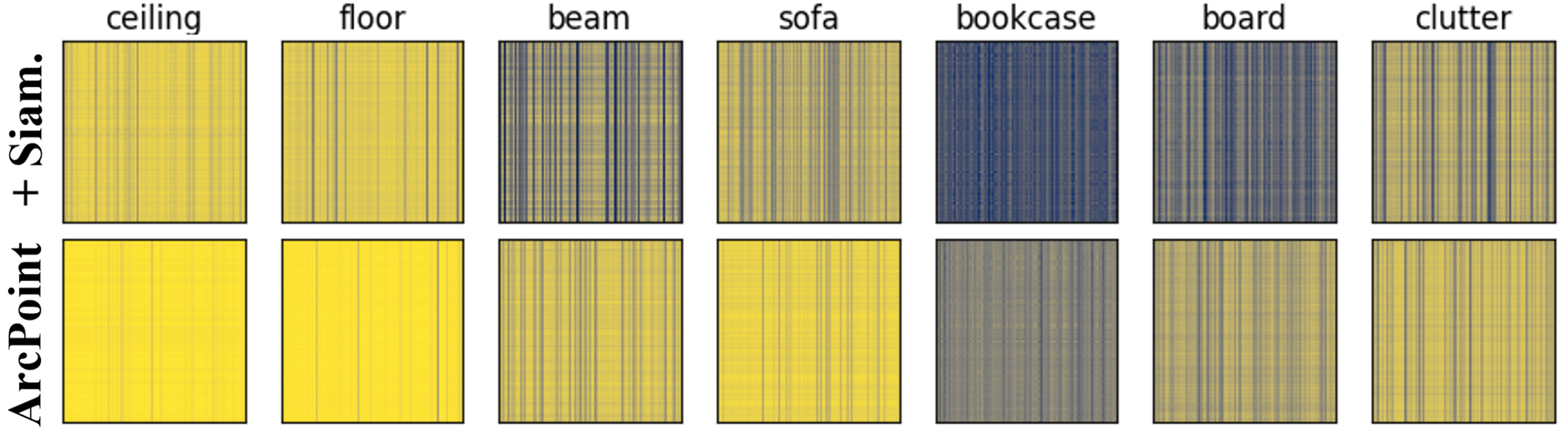}
\end{center}
\caption{Comparison of cosine similarity. 50,000 anchors and unannotated points with high entropy were randomly sampled from the S3DIS dataset. In each heatmap, the rows and columns indicate anchors and unannotated points, respectively.}
\label{cos_theta}
\end{figure}
Embedding on the hypersphere (e.g., ArcFace and ArcPoint) exhibited better performance compared with conventional losses. However, compared to ArcPoint loss, the gain of ArcFace was inevitably low because ArcFace could not be utilized in the optimization of the unlabeled points.
\input{table/ablation_gamma}
In contrast, ArcPoint achieved improvement (i.e., 1.1\%p and 2.5\%p, respectively) by applying an additive angular margin to the unlabeled points containing high entropy. To verify the effectiveness of ArcPoint, we visualized the cosine similarities between the anchors and unannotated points. Excluding the entropy block from the GaIA, the network was compared with the baseline including the Siamese branch. In Fig. \ref{cos_theta}, dark-colored vertical lines indicate that the unlabeled points have low cosine similarity compared to other anchors. That is, the ArcPoint loss effectively optimizes the unlabeled points by employing both anchors and angular margin penalty.

\subsection{Effectiveness of optimization with selective penalization}
To validate the selective penalization for the optimization effect, which focuses on points with high entropy, $F(\gamma)$ values were experimented with in multiple ranges. In Tab. \ref{ablation_gamma}, it is observed that the more attention imposed on the points containing high entropy ($F(0.5$ to $0.9) \uparrow$), the higher the performance is compared with applying the penalty to the point with low entropy ($F(0.1$ to $0.3) \downarrow$). In particular, although all points including high entropy were equally treated in the penalization ($F(0) \uparrow$), the performance was reduced. This is because the points containing high entropy were considered under the same conditions, not selectively penalized. This tendency was consistently observed in both 1pt and 1\% supervisions. In other words, it is effective to optimize the points, including high entropy, through selective penalization.


\begin{figure}[ht]
\begin{center}
\includegraphics[scale=0.233]{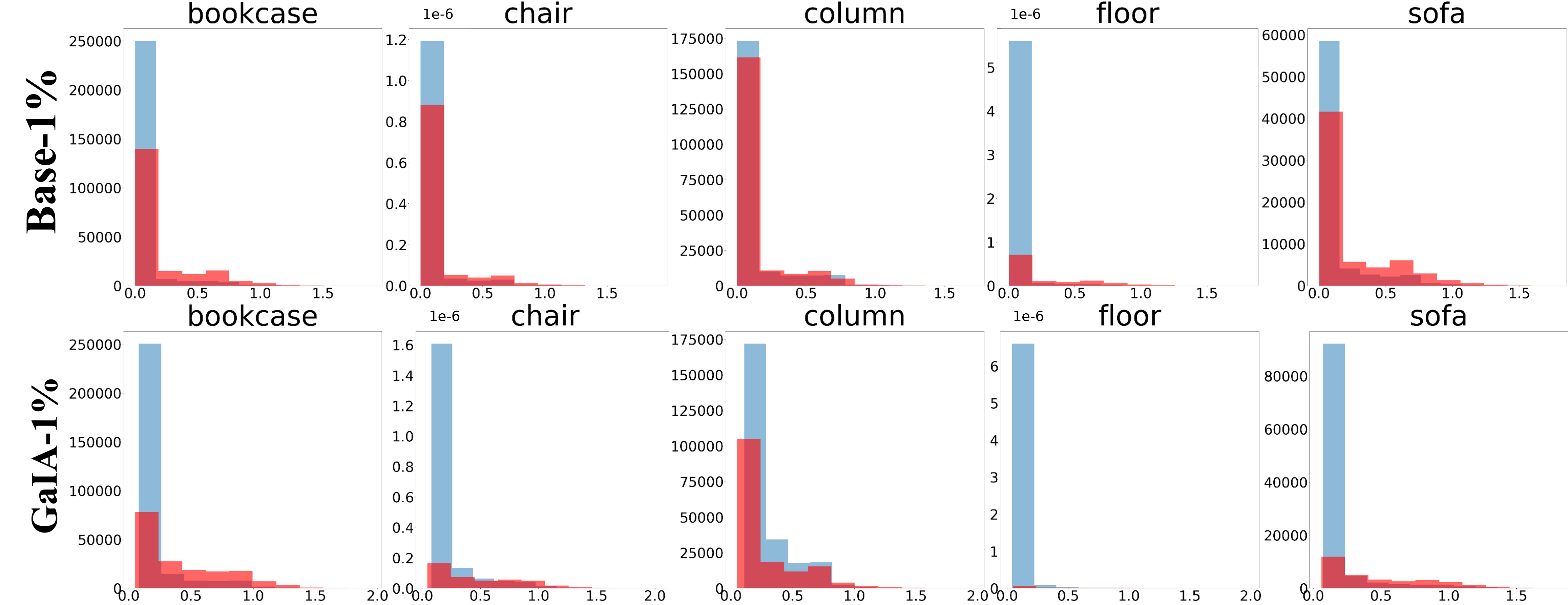}
\end{center}
\caption{Comparison of entropy distribution with respect to prediction. The x and y axes indicate entropy and the number of samples, respectively. Distribution highlighted with red indicates distribution of false prediction.}
\label{entropy_dist}
\end{figure}

\begin{figure}[ht]
\begin{center}
\includegraphics[scale=0.55]{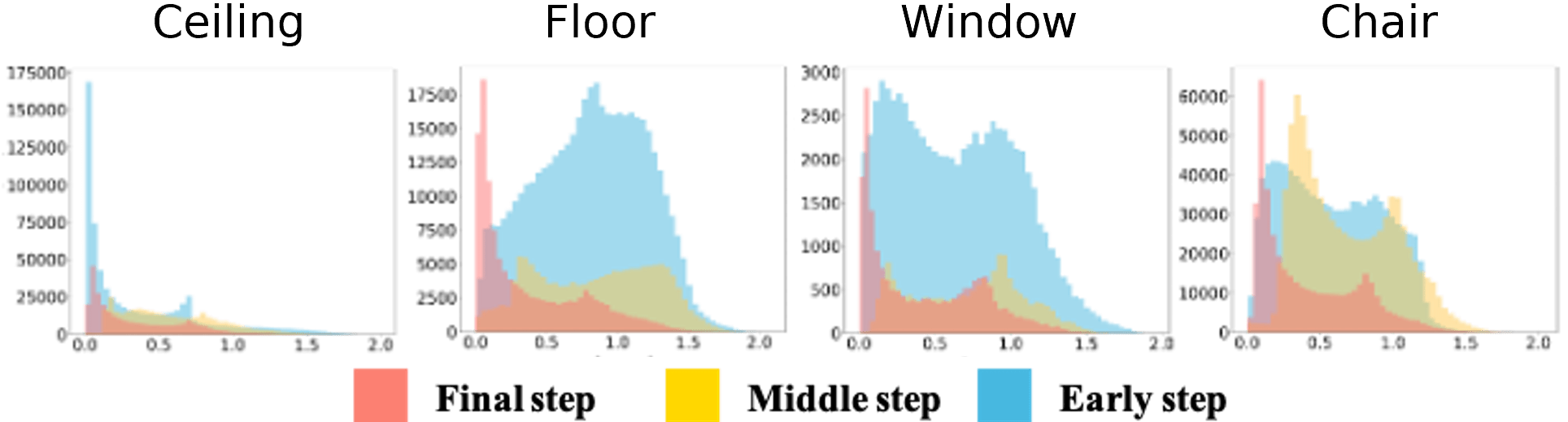}
\end{center}
\caption{Comparison of point-wise entropy variation for false predictions during training steps. X-axis indicates the entropy.}
\label{false_pred}
\end{figure}

\section{Discussion}
This study aims to reduce epistemic uncertainty by using the entropy of each point for effective and precise point cloud semantic segmentation. This claim includes the premise that the lower entropy of the network, the more precise the semantic segmentation result is. However, although the network contains low epistemic uncertainty, it can still estimate incorrectly. Hence, the entropy distribution was examined for each class and the distributions of true and false predictions were compared. In Fig. \ref{entropy_dist}, it is observed that GaIA alleviated the number of false predictions with low entropy compared with the baseline which included the Siamese branch. In particular, GaIA reduced the false predictions for the classes floor and chair, by approximately 6 M and 0.6 M, respectively. Moreover, we observed the number of false predictions progressively decreased along with epistemic uncertainty reduction following the training steps, as depicted in Fig. \ref{false_pred}. In other words, GaIA, which alleviates epistemic uncertainty, resulted in the reduction of false prediction with high reliability. Further analysis on the false prediction with high reliability is offered in Supplementary.

\section{Conclusion}
This study addressed epistemic uncertainty reduction in effective and precise point cloud semantic segmentation. Graphical information gain based attention network called GaIA was proposed. The graphical information gain and anchor-based additive angular margin loss called ArcPoint were main contributions of our approach. Specifically, the graphical information gain represents the reliable information by computing the relative entropy between the entropy of the target point and that of its neighborhoods. ArcPoint effectively optimizes unlabeled points containing high entropy. The experimental results of our method on two large-scale datasets demonstrated the improved performance of the proposed method compared with existing weakly supervised point cloud semantic segmentation methods.

\section*{Acknowledgements} This research was supported by Brain Korea 21 FOUR.

\section*{Supplementary}
The remainder of the supplementary materials are organized as follows. Section \ref{GI} explains ablation studies related to graphical information gain. Section \ref{arcpoint} describes further analysis of the ArcPoint loss function. Section \ref{trade_off} presents a trade-off between epistemic uncertainty reduction and false prediction with high reliability. Finally, Section \ref{s3dis_viz_comparison} compares the S3DIS visualization results of Baseline and GaIA.

\section{Analysis on graphical information gain}\label{GI}
\subsection{Neighborhood aggregation based on Euclidean distance}
We conducted ablation studies to investigate the effect of Euclidean distance-based neighbor aggregation in the entropy block. Experiments were performed on a network with a Siamese branch and ArcPoint. As listed in lines 1 and 2 of Tab. \ref{sup_ablation_EB}, we compared the original entropy aggregation $\widetilde{H}=\sum_k{H_k}$ with the aggregation, which is inversely proportional to the Euclidean distance of the neighborhood, $\widetilde{H} = \sum_{k}{(D_{k})^{-2}} \cdot H_{k} / \sum_j{(D_{k})^{-2}}$. When we calibrated $H_{k}$ based on the Euclidean distance, the performance surpassed that of only employing the original entropy by 2.2\%p.

\subsection{Effectiveness of neighbor aggregation}
To investigate which information is more important, either reliable point-wise attention or reliable neighbor information for updating the point representation, ablation studies were conducted. Lines 4 and 5 of Tab. \ref{sup_ablation_EB} show that the neighborhood aggregation $X = X + (X \otimes GI) + X^N$ outperformed the reliable point-wise attention $X = X + (X \otimes GI)$. Here, aggregation without normalization showed unsatisfactory performance compared with the point-wise attention $X = X + (X \otimes GI)$ because of the overfitting problem. In other words, normalized neighbor aggregation was effective in updating the uncertain point representation compared with only enhancing the reliable points.

\subsection{Entropy block organization}
Applying the entropy block to each decoder block results in computational inefficiency. Therefore, we exclude the entropy block from the decoder in our network. When we organized both the encoder and decoder with the entropy block, the inefficiency increased compared with the performance gain.
\input{table/sup_ablation_EB}
\input{table/EB_framework}
In line 2 of Tab. \ref{EB_loss}, the training time (seconds per iteration) increased by 1.5 seconds while the performance improved 0.1\%p at 1\% on S3DIS dataset. With a similar tendency, for the ScanNet-v2 dataset, the seconds per iteration increased by 1.9 seconds in comparison with the performance gain of 0.2\%p. Based on the experimental results, we excluded the entropy block from the decoder. This means that obtaining high-quality feature representations contributes significantly to performance gain.

\subsection{Comparison of entropy reduction}
\begin{figure*}[h]
\begin{center}
\includegraphics[scale=0.185]{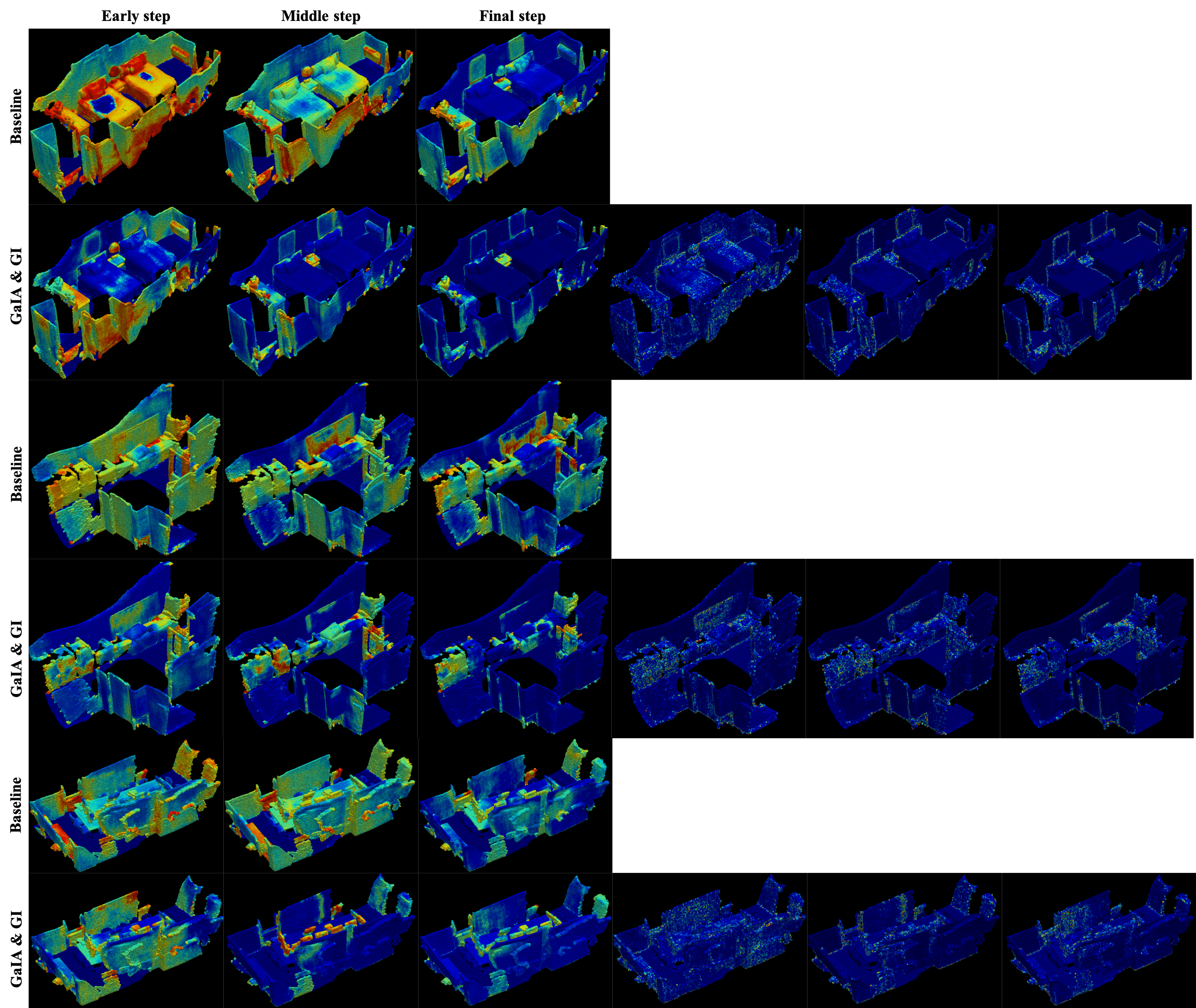}
\end{center}
\caption{Comparison of entropy variation.}
\label{entropy_decrease}
\end{figure*}

GaIA was compared with the baseline network by visualizing entropy to investigate the entropy reduction effect. As depicted in Fig. \ref{entropy_decrease}, we observed that GaIA more broadly alleviated the entropy overall training steps compared to the baseline. This is because $GI$ contributed to updating the uncertain points near the ambiguous decision boundary of the network toward the semantically similar points. As mentioned in Section 3.2 from the original study, $GI$ discriminates the relative entropy between the entropy of the target point and that of its neighbors to identify reliable information.

\begin{figure*}[ht]
\begin{center}
\includegraphics[scale=.425]{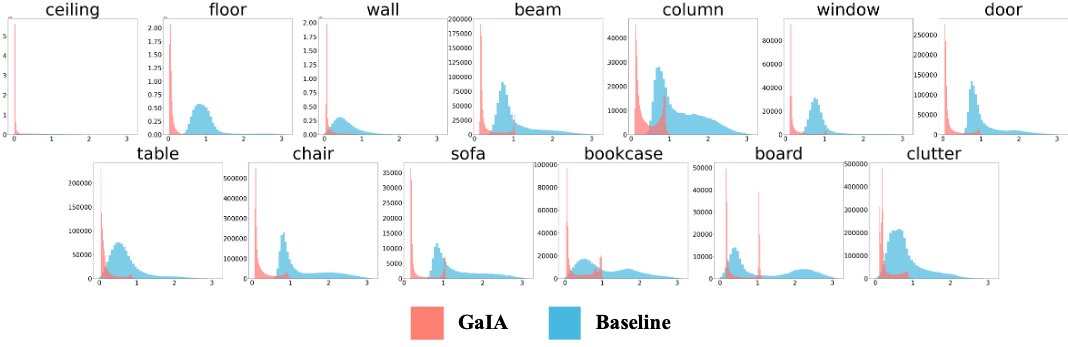}
\end{center}
\caption{Comparison with ArcPoint and conventional softmax loss under the 1\% annotation. The x-axis indicates the angle between the class-wise prototypical weight matrix $W_y$ and the unannotated point.}
\label{comparison_theta}
\end{figure*}

\begin{figure*}[ht]
\begin{center}
\includegraphics[scale=0.45]{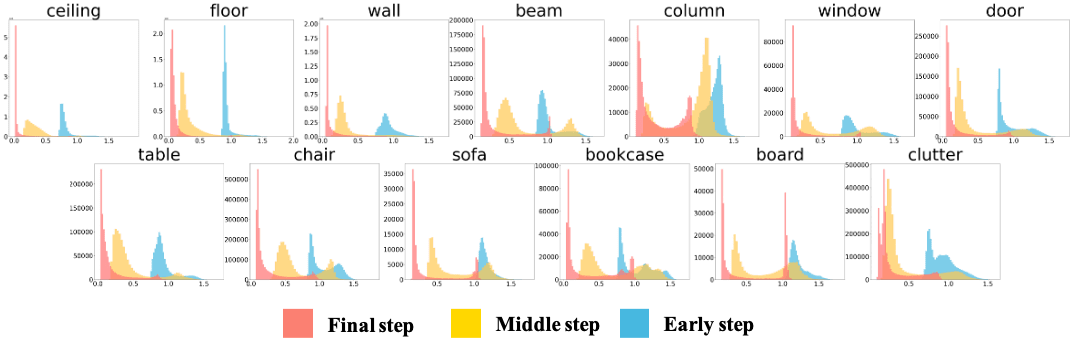}
\end{center}
\caption{Comparison of angle variation corresponding to training steps under the 1\% annotation. The x-axis indicates the angle between the class-wise prototypical weight matrix $W_y$ and the unannotated point.}
\label{arcpoint_theta}
\end{figure*}

\begin{figure*}[ht]
\begin{center}
\includegraphics[scale=0.5]{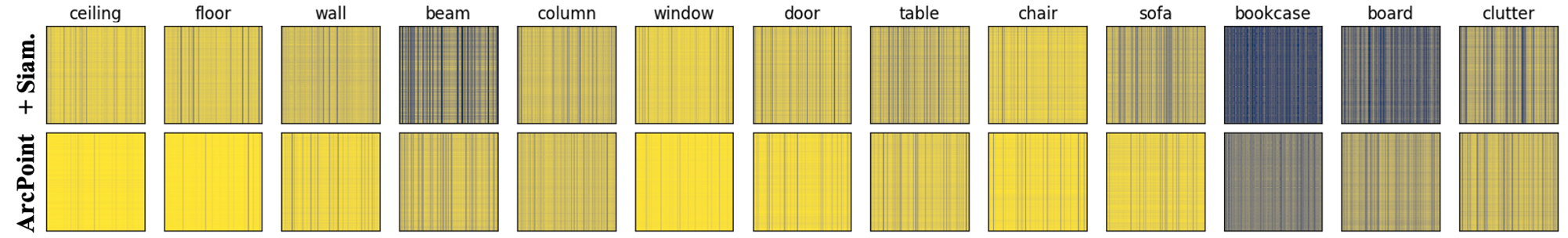}
\end{center}
\caption{Comparison of cosine similarity. 50,000 anchors and unannotated points with high entropy were randomly sampled from the S3DIS dataset. In each heatmap, the rows and columns indicate anchors and unannotated points, respectively.}
\label{loss_effect}
\end{figure*}

\begin{figure*}[h]
\begin{center}
\includegraphics[scale=0.075]{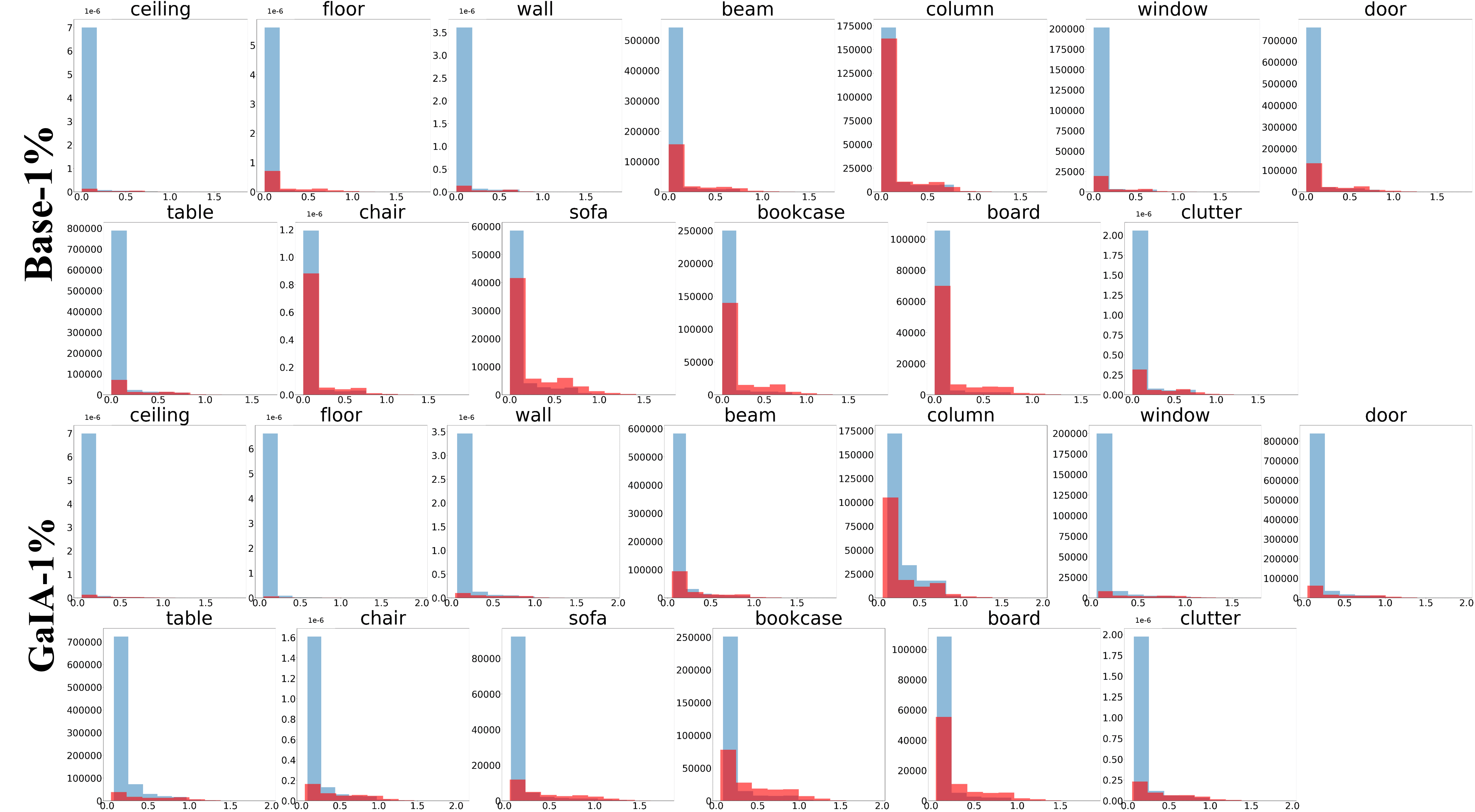}
\end{center}
\caption{Comparison of entropy distribution with respect to prediction. The x and y axes indicate entropy and the number of samples, respectively. Distribution highlighted with red indicates distribution of false prediction.}
\label{entropy_dist}
\end{figure*}

\section{Analysis on ArcPoint loss}\label{arcpoint}
We addressed that ArcPoint loss optimizes unlabeled points toward semantically similar points by penalizing the unannotated points using an additive angular margin. In addition to the analysis in Sections 5.2 and 6 from the original study, we compared ArcPoint loss with the conventional softmax loss function by using the distribution of angle. The angle was measured using the class-wise prototypical weight matrix $W_y$ and the unlabeled point as follows: $\theta_y(x_j^u) = arccos(\frac{W_{y}^\intercal \cdot x_j^u}{||W_{y}|| \cdot ||x_j^u||})$. As depicted in Fig. 8 from the original study, for the baseline network employing softmax loss, the ratio of false predictions in the column, chair, bookcase, and board classes were higher compared with ArcPoint. Moreover, as shown in Fig. \ref{comparison_theta}, the angle distribution of the baseline network commonly exhibited a long tail distribution for the mentioned classes. This is because the baseline could not enhance both intra-class density and inter-class distinction. In contrast, ArcPoint showed a short-tail distribution, which indicates that ArcPoint effectively optimized the unlabeled points for all classes compared to the existing softmax loss. To observe angle variation corresponding to training steps, the distributions of angle were compared. As depicted in Fig. \ref{arcpoint_theta}, the average level of distance, which was measured by $\theta_y(x_j^u)$, gradually decreased during the training. To verify the effectiveness of ArcPoint, we visualized the cosine similarities between the anchors and unannotated points in Fig. \ref{loss_effect}.

\section{Trade-off between epistemic uncertainty reduction and false prediction with high reliability}\label{trade_off}
\subsection{Reduction of false predictions}
As mentioned in Section 6 of the original study, although the network contains low epistemic uncertainty, it can still estimate incorrectly. Hence, the entropy distribution was examined for each class and the distributions of true and false predictions were compared. As depicted in Fig. \ref{entropy_dist}, it is observed that GaIA alleviated the number of false predictions with low entropy compared with the baseline which included the Siamese branch. False predictions with high reliability were reduced for all classes.

\subsection{Point-wise entropy variation}
\begin{figure*}[ht]
\begin{center}
\includegraphics[scale=0.44]{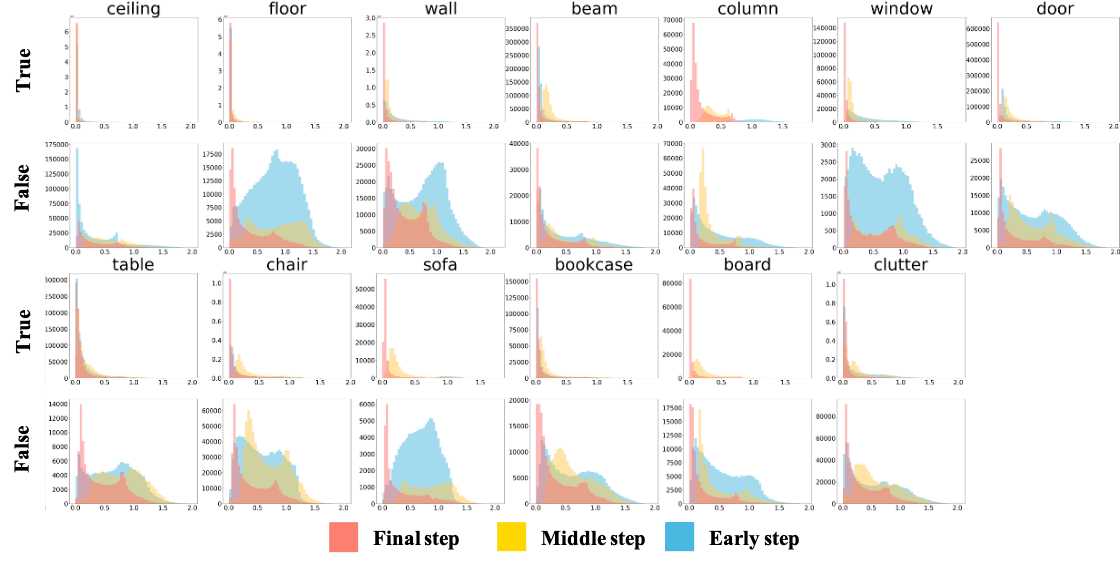}
\end{center}
\caption{Comparison of point-wise entropy variation corresponding to prediction and training steps under the 1\% annotation. The x-axis indicates the entropy.}
\label{entropy_trade_off}
\end{figure*}

\begin{figure*}[ht]
\begin{center}
\includegraphics[scale=0.44]{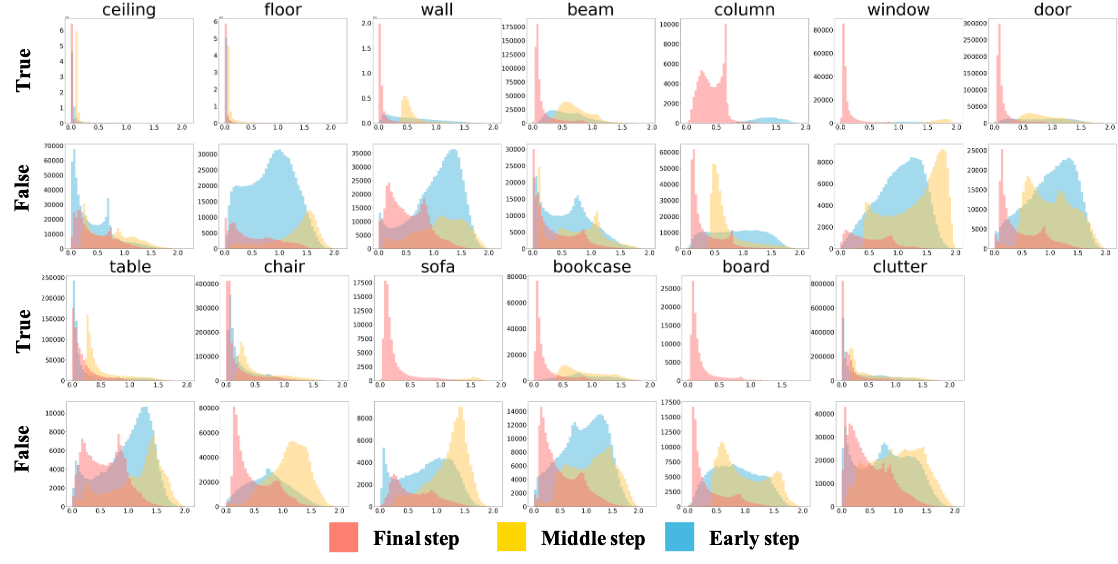}
\end{center}
\caption{Comparison of point-wise entropy variation corresponding to prediction and training steps under the 1pt annotation. The x-axis indicates the entropy.}
\label{entropy_trade_off_1pt}
\end{figure*}

Based on the experimental results in Fig. \ref{comparison_theta} and \ref{arcpoint_theta}, a few unannotated points (in the column, sofa, bookcase, and board classes) could not be optimized well despite further training steps. Indeed, for these classes, the more training the network, the closer the distribution to bimodal is. To examine this optimization discrepancy, we visualized the point-wise entropy variation corresponding to both network prediction and training steps. In Fig. \ref{entropy_trade_off}, we observed the number of false predictions progressively decreased along with epistemic uncertainty reduction following the training steps. Although the uncertainty reduction significantly improved the semantic segmentation performance compared with the baseline, it was confirmed that the uncertainty reduction resulted in a trade-off that false predictions with high reliability increased under the 1\% annotation. To investigate whether this trade-off is consistently observed under the sparser annotation, we visualized the entropy distribution under the 1pt annotation. Similar to the 1\% annotation, Fig. \ref{entropy_trade_off_1pt} exhibited a ratio of false prediction and the average level of entropy increased.

\input{table/sup_ablation_false_prediction}

\subsection{Consistency for false prediction with high reliability}
Based on these observations, we assumed that the points, which were inaccurately predicted with high reliability in the early training step, consistently had low entropy. To validate this assumption, we examined the consistency of false predictions with high reliability. As listed in Tab. \ref{sup_false_prediction}, the number of false predictions for each class was measured corresponding to the training steps. Consistency was observed in the column, sofa, bookcase, and board classes, which exhibited a trade-off. The ratio of false predictions with high reliability for the mentioned classes commonly occupied higher levels over the training steps compared with other classes.

\section{Comparison of qualitative results on S3DIS}\label{s3dis_viz_comparison}
\begin{figure*}[ht]
\begin{center}
\includegraphics[scale=0.285]{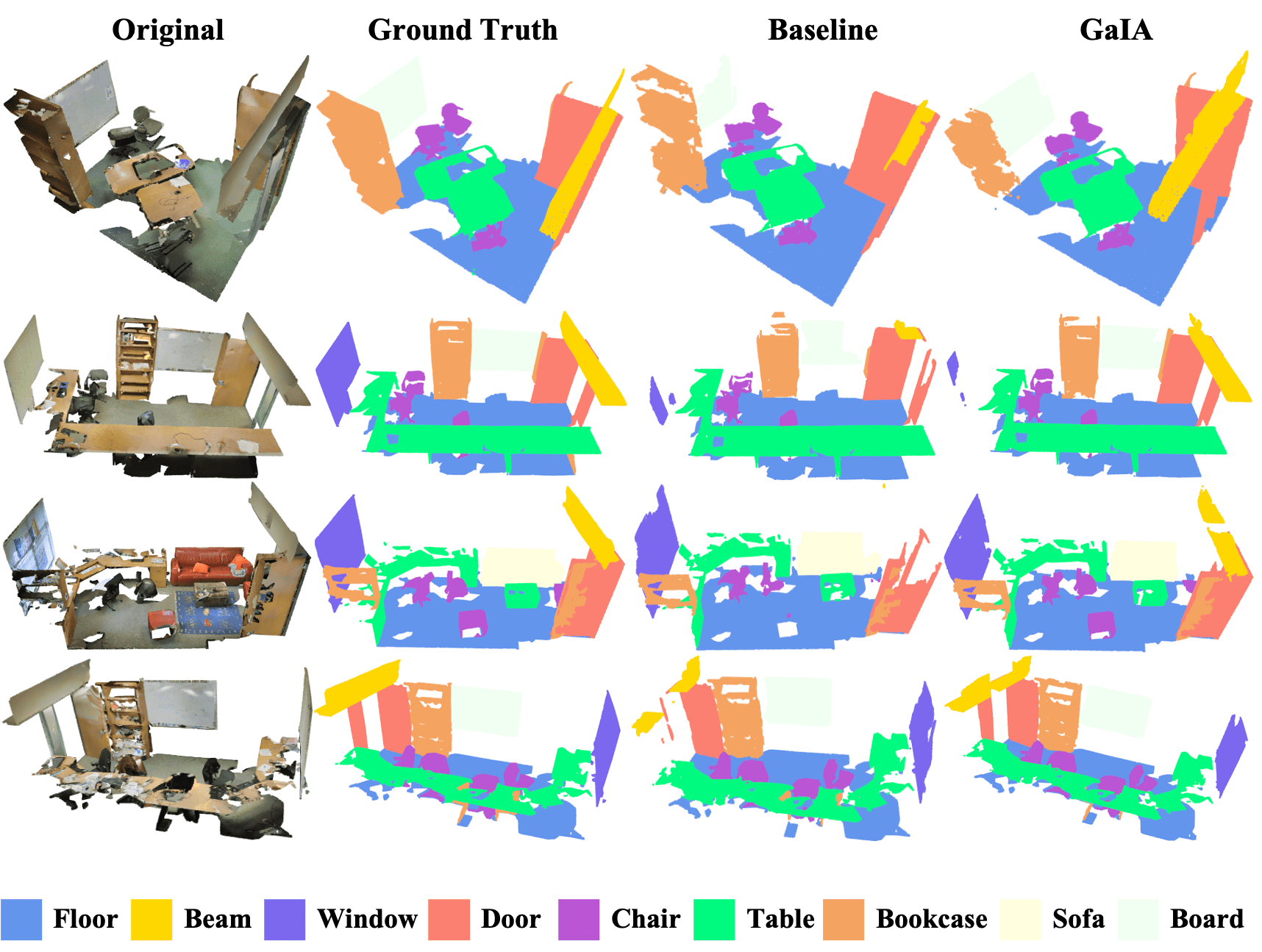}
\end{center}
  \caption{Comparison of qualitative results on S3DIS.}
\label{s3dis_viz}
\end{figure*}

As depicted in Fig. \ref{s3dis_viz}, GaIA performed better, which is closer to Ground Truth, than the baseline.

{\small
\bibliographystyle{ieee_fullname}
\bibliography{ref}
}

\end{document}

%% file: table/benchmark_s3dis.tex
\begin{table}[ht]
\centering
\caption{Comparison with existing methods on S3DIS dataset.}
\resizebox{0.45\textwidth}{!}{%
{\Large
\label{benchmark_s3dis}
\begin{tabular}{c|c|c|c}
\hline
Method & Supervision & Area 5 & 6-Fold\\
\hline
PointNet \cite{qi2017pointnet} & 100\% & 41.1 & 47.6\\
PointNet++ \cite{qi2017pointnet++} & 100\% & -- & 54.5\\ 
PointCNN \cite{Li2018PointCNNCO}      & 100\% & 57.3 & 65.4\\ 
KPConv \cite{thomas2019KPConv} & 100\% & 67.1 & 70.6\\ 
MinkowskiNet \cite{choy20194d} & 100\% & 65.3 & --\\
RandLA-Net \cite{hu2020randla} & 100\% & 63.0 & 70.0\\ 
PointASNL \cite{yan2020pointasnl} & 100\% & -- & 68.7\\ 
PointTransformer \cite{zhao2021point}& 100\% & 70.4 & \bf 73.5\\ 
Hou \etal \cite{hou2021exploring} & 100\% & \bf 72.2 & --\\
CBL \cite{tang2022contrastive} & 100\% & 69.4 & 73.1 \\
HybridCR \cite{Li_2022_CVPR} & 100\% & 65.8 & 70.7\\
\hline

Zhang \etal \cite{zhang2021weakly} & 1\% & 61.8 & 65.9\\ 
PSD \cite{zhang2021perturbed} & 1\% & 63.5 & 68.0\\ 
HybridCR \cite{Li_2022_CVPR} & 1\% & 65.3 & 69.2\\
GaIA (Ours) & 1\% & \bf 66.5 & \bf 70.8\\
\hline
Xu and Lee \cite{xu2020weakly} & 1pt (0.2\%) & 44.5 & --\\
PSD \cite{zhang2021perturbed} & 1pt (0.03\%) & 48.2 & --\\ 
HybridCR \cite{Li_2022_CVPR} & 1pt (0.03\%) & 51.5 & --\\
OTOC \cite{liu2021one} & 1pt (0.02\%) & 43.7 & --\\
MIL Transformer \cite{Yang_2022_CVPR} & 1pt (0.02\%) & 51.4 & --\\
GaIA (Ours) & 1pt (0.02\%) & \bf 53.7 & --\\
\hline
\end{tabular}}}
\end{table}

%% file: table/benchmark_scannet.tex
\begin{table}[ht]
\centering
\caption{Comparison with existing methods on ScanNet-v2.}
\resizebox{0.45\textwidth}{!}{%
{\Large
\label{benchmark_scannet}
\begin{tabular}{c|c|c}
\hline
Method & Supervision & mIoU\\
\hline
PointNet++ \cite{qi2017pointnet++} & 100\% & 33.9\\ 
PointCNN \cite{Li2018PointCNNCO}      & 100\% & 45.8\\ 
KPConv \cite{thomas2019KPConv} & 100\% & 68.4\\ 

RandLA-Net \cite{hu2020randla} & 100\% & 64.5\\ 
PointASNL \cite{yan2020pointasnl} & 100\% & 66.6\\ 
MinkowskiNet \cite{choy20194d} & 100\% & 73.6\\
VMNet \cite{hu2021vmnet} & 100\% & 74.6\\ 
BPNet \cite{hu2021bidirectional} & 100\% & 74.9\\ 
Mix3D \cite{nekrasov2021mix3d} & 100\% & \bf 78.1 \\
\hline
Zhang \etal \cite{zhang2021weakly} & 1\% & 51.1\\ 
PSD \cite{zhang2021perturbed} & 1\% & 54.7\\ 
HybridCR \cite{Li_2022_CVPR} & 1\% & 56.8\\ 
GaIA (Ours) & 1\% & \bf 65.2\\
\hline
Hou \etal \cite{hou2021exploring} & 20 pts / scene & 55.5\\
OTOC \cite{liu2021one} & 20 pts / scene & 59.4\\
MIL Transformer \cite{Yang_2022_CVPR} & 20 pts / scene & 54.4\\
GaIA (Ours) & 20 pts / scene & \bf 63.8\\
\hline
GaIA (Ours) & avg 7.8 pts / scene (1pt) & 52.1\\
\hline
\end{tabular}}}
\end{table}

%% file: table/ablation_module.tex
\begin{table}[ht]
\centering
\caption{Comparison of the quantitative results on ScanNet-v2 validation set corresponding to the proposed components. ($\cdot$) indicates officially measured test scores.}
\resizebox{0.46\textwidth}{!}{%
{\Large
\label{table_ablation_module}
\begin{tabular}{ccccc|cc}
\hline
Base. & Sia. & EB. & AP. & AF. & 1pt (0.005\%) & 1\%\\
\hline
\checkmark & & & & & 33.2 (43.6) & 42.7 (53.9)\\
\checkmark & \checkmark & & & & 37.4 & 49.5\\
\checkmark & \checkmark & & \checkmark & & 39.1 & 51.7\\
\checkmark & \checkmark & \checkmark & & & 40.8 & 52.4\\
\checkmark & \checkmark & \checkmark & & \checkmark & 41.1 & 52.8\\
\checkmark &\checkmark & \checkmark & \checkmark & & 41.9 ({\bf 52.1})& 54.9 ({\bf 65.2})\\
\hline
\end{tabular}}}
\end{table}

%% file: table/ablation_gamma.tex
\begin{table}[ht]
\centering
\caption{Comparison of the performance on ScanNet-v2 validation set corresponding to $F(\gamma)$. $F(\gamma)$ and ($\cdot$) denote the $\gamma$ quantile of $H$ and officially tested score, respectively.}
\resizebox{0.47\textwidth}{!}{%
{\Large
\label{ablation_gamma}
\begin{tabular}{c|c|ccc|ccc}
\hline
Sup. & $0 \uparrow$ & $0.1 \downarrow$ & $0.3 \downarrow$ & $0.5 \downarrow$ & $0.5 \uparrow$ & $0.7 \uparrow$ & $F(0.9) \uparrow$\\
\hline
1\% & 52.1 & 49.3 (59.5) & 49.5 & 49.8 & 53.0 & 53.8 & \bf 54.9 (65.2)\\
1pt & 39.2 & 37.1 (47.4) & 37.7 & 38.1 & 40.6 & 41.1 & \bf 41.9 (52.1)\\
\hline
\end{tabular}}}
\end{table}

%% file: table/sup_ablation_EB.tex
\begin{table}[ht]
\centering
\caption{Comparison of components applied in entropy block on ScanNet-v2. ($\cdot$) and $*$ denote the officially measured test scores and excluded neighborhood normalization, respectively.}
\label{sup_ablation_EB}
\begin{tabular}{cccc|cc}
\hline
$H_k$ & $(D_k^{-2} \cdot H_k)$ & $GI$ & $X^N$ & 1\%\\
\hline
\checkmark & & \checkmark & \checkmark & 52.7\\
\checkmark & \checkmark & \checkmark & \checkmark & 54.9 (65.2)\\
\hline
\checkmark & \checkmark & & \checkmark & 52.1$^*$\\
\checkmark & \checkmark & \checkmark & & 52.8\\
\checkmark & \checkmark & & \checkmark & 53.6\\
\checkmark & \checkmark & \checkmark & \checkmark & 54.9 (65.2)\\

\hline
\end{tabular}
\end{table}

%% file: table/EB_framework.tex
\begin{table}[ht]
\centering
\caption{Comparison of applying entropy block to GaIA framework. The results of ScanNet-v2 are measured by validation set.}
\label{EB_loss}
\begin{tabular}{c|cc|c|c}
\hline
Dataset & Enc. & Dec. & Sec/iter & 1\%\\
\hline
\multirow{2}{*}{S3DIS} & \checkmark & & 2.4 sec & 66.5\\
& \checkmark & \checkmark & 3.9 sec & 66.8\\
\hline
\multirow{2}{*}{ScanNet-v2} & \checkmark & & 4.3 sec & 54.9\\
& \checkmark & \checkmark & 6.2 sec & 55.3\\
\hline
\end{tabular}
\end{table}

%% file: table/sup_ablation_false_prediction.tex
\begin{table*}[ht]
\centering
\caption{Comparison of the number of false predictions corresponding to training steps.}
\label{sup_false_prediction}
\begin{tabular}{c|ccc|c|c|c}
\hline
Class & Early & Mid & Final & Early \& Mid & Mid \& Final & Early \& Mid \& Final\\
\hline
Ceiling & 662,251 & 380,898 & 281,615 & 344,494 & 197,691 & 193,557\\
Floor & 456,441 & 413,410 & 122,183 & 295,146 & 98,293 & 93,063\\
Wall & 637,130 & 418,767 & 286,331 & 274,121 & 138,686 & 108,794\\
Beam & 193,450 & 252,106 & 151,278 & 164,453 & 121,664 & 100,832\\
Column & 365,368 & 315,687 & {\bf 154,541} & 303,338 & 143,604 & {\bf 139,579}\\
Window & 75,755 & 107,105 & 19,849 & 60,126 & 18,173 & 15,699\\
Door & 370,788 & 291,378 & 121,271 & 211,820 & 75,458 & 59,609\\
Table & 154,154 & 159,808 & 107,214 & 113,659 & 78,010 & 71,358\\
Chair & 1,031,409 & 752,985 & 413,881 & 629,131 & 279,760 & 244,589\\
Sofa & 115,168 & 60,429 & {\bf 30,085} & 58,586 & 25,965 & {\bf 25,787}\\
Bookcase & 222,362 & 210,304 & {\bf 168,781} & 186,488 & 149,283 & {\bf 144,135}\\
Board & 205,920 & 110,817 & {\bf 83,586} & 110,592 & 74,966 & {\bf 74,833}\\
Clutter & 680,719 & 625,378 & 533,408 & 403,932 & 321,078 & 250,165\\
\hline
\end{tabular}
\end{table*}

%% file: main.bbl
\begin{thebibliography}{10}\itemsep=-1pt

\bibitem{amodei2016concrete}
Dario Amodei, Chris Olah, Jacob Steinhardt, Paul Christiano, John Schulman, and
  Dan Man{\'e}.
\newblock Concrete problems in ai safety.
\newblock {\em arXiv preprint arXiv:1606.06565}, 2016.

\bibitem{armeni20163d}
Iro Armeni, Ozan Sener, Amir~R Zamir, Helen Jiang, Ioannis Brilakis, Martin
  Fischer, and Silvio Savarese.
\newblock 3d semantic parsing of large-scale indoor spaces.
\newblock In {\em Proceedings of the IEEE Conference on Computer Vision and
  Pattern Recognition}, pages 1534--1543, 2016.

\bibitem{bromley1993signature}
Jane Bromley, James~W Bentz, L{\'e}on Bottou, Isabelle Guyon, Yann LeCun, Cliff
  Moore, Eduard S{\"a}ckinger, and Roopak Shah.
\newblock Signature verification using a “siamese” time delay neural
  network.
\newblock {\em International Journal of Pattern Recognition and Artificial
  Intelligence}, 7(04):669--688, 1993.

\bibitem{chang2015shapenet}
Angel~X Chang, Thomas Funkhouser, Leonidas Guibas, Pat Hanrahan, Qixing Huang,
  Zimo Li, Silvio Savarese, Manolis Savva, Shuran Song, Hao Su, et~al.
\newblock Shapenet: An information-rich 3d model repository.
\newblock {\em arXiv preprint arXiv:1512.03012}, 2015.

\bibitem{cheng2021sspc}
Mingmei Cheng, Le Hui, Jin Xie, and Jian Yang.
\newblock Sspc-net: Semi-supervised semantic 3d point cloud segmentation
  network.
\newblock In {\em Proceedings of the AAAI Conference on Artificial
  Intelligence}, volume~35, pages 1140--1147, 2021.

\bibitem{choi2019gaussian}
Jiwoong Choi, Dayoung Chun, Hyun Kim, and Hyuk-Jae Lee.
\newblock Gaussian yolov3: An accurate and fast object detector using
  localization uncertainty for autonomous driving.
\newblock In {\em Proceedings of the IEEE/CVF International Conference on
  Computer Vision}, pages 502--511, 2019.

\bibitem{choy20194d}
Christopher Choy, JunYoung Gwak, and Silvio Savarese.
\newblock 4d spatio-temporal convnets: Minkowski convolutional neural networks.
\newblock In {\em Proceedings of the IEEE/CVF Conference on Computer Vision and
  Pattern Recognition}, pages 3075--3084, 2019.

\bibitem{dai2017scannet}
Angela Dai, Angel~X Chang, Manolis Savva, Maciej Halber, Thomas Funkhouser, and
  Matthias Nie{\ss}ner.
\newblock Scannet: Richly-annotated 3d reconstructions of indoor scenes.
\newblock In {\em Proceedings of the IEEE conference on computer vision and
  pattern recognition}, pages 5828--5839, 2017.

\bibitem{deng2019arcface}
Jiankang Deng, Jia Guo, Niannan Xue, and Stefanos Zafeiriou.
\newblock Arcface: Additive angular margin loss for deep face recognition.
\newblock In {\em Proceedings of the IEEE/CVF Conference on Computer Vision and
  Pattern Recognition}, pages 4690--4699, 2019.

\bibitem{engelmann2017exploring}
Francis Engelmann, Theodora Kontogianni, Alexander Hermans, and Bastian Leibe.
\newblock Exploring spatial context for 3d semantic segmentation of point
  clouds.
\newblock In {\em Proceedings of the IEEE international conference on computer
  vision workshops}, pages 716--724, 2017.

\bibitem{feng2018towards}
Di Feng, Lars Rosenbaum, and Klaus Dietmayer.
\newblock Towards safe autonomous driving: Capture uncertainty in the deep
  neural network for lidar 3d vehicle detection.
\newblock In {\em 2018 21st International Conference on Intelligent
  Transportation Systems (ITSC)}, pages 3266--3273. IEEE, 2018.

\bibitem{gal2016uncertainty}
Yarin Gal et~al.
\newblock Uncertainty in deep learning.
\newblock 2016.

\bibitem{graham20183d}
Benjamin Graham, Martin Engelcke, and Laurens Van Der~Maaten.
\newblock 3d semantic segmentation with submanifold sparse convolutional
  networks.
\newblock In {\em Proceedings of the IEEE conference on computer vision and
  pattern recognition}, pages 9224--9232, 2018.

\bibitem{harpur1997low}
George~Francis Harpur.
\newblock {\em Low entropy coding with unsupervised neural networks}.
\newblock PhD thesis, Citeseer, 1997.

\bibitem{harpur1996development}
George~F Harpur and Richard~W Prager.
\newblock Development of low entropy coding in a recurrent network.
\newblock {\em Network: computation in neural systems}, 7(2):277--284, 1996.

\bibitem{hou2021exploring}
Ji Hou, Benjamin Graham, Matthias Nie{\ss}ner, and Saining Xie.
\newblock Exploring data-efficient 3d scene understanding with contrastive
  scene contexts.
\newblock In {\em Proceedings of the IEEE/CVF Conference on Computer Vision and
  Pattern Recognition}, pages 15587--15597, 2021.

\bibitem{hu2020randla}
Qingyong Hu, Bo Yang, Linhai Xie, Stefano Rosa, Yulan Guo, Zhihua Wang, Niki
  Trigoni, and Andrew Markham.
\newblock Randla-net: Efficient semantic segmentation of large-scale point
  clouds.
\newblock In {\em Proceedings of the IEEE/CVF Conference on Computer Vision and
  Pattern Recognition}, pages 11108--11117, 2020.

\bibitem{hu2021bidirectional}
Wenbo Hu, Hengshuang Zhao, Li Jiang, Jiaya Jia, and Tien-Tsin Wong.
\newblock Bidirectional projection network for cross dimension scene
  understanding.
\newblock In {\em Proceedings of the IEEE/CVF Conference on Computer Vision and
  Pattern Recognition}, pages 14373--14382, 2021.

\bibitem{hu2021vmnet}
Zeyu Hu, Xuyang Bai, Jiaxiang Shang, Runze Zhang, Jiayu Dong, Xin Wang,
  Guangyuan Sun, Hongbo Fu, and Chiew-Lan Tai.
\newblock Vmnet: Voxel-mesh network for geodesic-aware 3d semantic
  segmentation.
\newblock In {\em Proceedings of the IEEE/CVF International Conference on
  Computer Vision}, pages 15488--15498, 2021.

\bibitem{jiang2020pointgroup}
Li Jiang, Hengshuang Zhao, Shaoshuai Shi, Shu Liu, Chi-Wing Fu, and Jiaya Jia.
\newblock Pointgroup: Dual-set point grouping for 3d instance segmentation.
\newblock In {\em Proceedings of the IEEE/CVF Conference on Computer Vision and
  Pattern Recognition}, pages 4867--4876, 2020.

\bibitem{kendall2017uncertainties}
Alex Kendall and Yarin Gal.
\newblock What uncertainties do we need in bayesian deep learning for computer
  vision?
\newblock {\em Advances in neural information processing systems}, 30, 2017.

\bibitem{koch2015siamese}
Gregory Koch, Richard Zemel, Ruslan Salakhutdinov, et~al.
\newblock Siamese neural networks for one-shot image recognition.
\newblock In {\em ICML deep learning workshop}, volume~2. Lille, 2015.

\bibitem{labonte2019we}
Tyler LaBonte, Carianne Martinez, and Scott~A Roberts.
\newblock We know where we don't know: 3d bayesian cnns for credible geometric
  uncertainty.
\newblock {\em arXiv preprint arXiv:1910.10793}, 2019.

\bibitem{Li_2022_CVPR}
Mengtian Li, Yuan Xie, Yunhang Shen, Bo Ke, Ruizhi Qiao, Bo Ren, Shaohui Lin,
  and Lizhuang Ma.
\newblock Hybridcr: Weakly-supervised 3d point cloud semantic segmentation via
  hybrid contrastive regularization.
\newblock In {\em Proceedings of the IEEE/CVF Conference on Computer Vision and
  Pattern Recognition (CVPR)}, pages 14930--14939, June 2022.

\bibitem{Li2018PointCNNCO}
Yangyan Li, Rui Bu, Mingchao Sun, Wei Wu, Xinhan Di, and Baoquan Chen.
\newblock Pointcnn: Convolution on x-transformed points.
\newblock In {\em NeurIPS}, 2018.

\bibitem{liu2021one}
Zhengzhe Liu, Xiaojuan Qi, and Chi-Wing Fu.
\newblock One thing one click: A self-training approach for weakly supervised
  3d semantic segmentation.
\newblock In {\em Proceedings of the IEEE/CVF Conference on Computer Vision and
  Pattern Recognition}, pages 1726--1736, 2021.

\bibitem{malinin2018predictive}
Andrey Malinin and Mark Gales.
\newblock Predictive uncertainty estimation via prior networks.
\newblock {\em Advances in neural information processing systems}, 31, 2018.

\bibitem{mo2019partnet}
Kaichun Mo, Shilin Zhu, Angel~X Chang, Li Yi, Subarna Tripathi, Leonidas~J
  Guibas, and Hao Su.
\newblock Partnet: A large-scale benchmark for fine-grained and hierarchical
  part-level 3d object understanding.
\newblock In {\em Proceedings of the IEEE/CVF conference on computer vision and
  pattern recognition}, pages 909--918, 2019.

\bibitem{nair2020exploring}
Tanya Nair, Doina Precup, Douglas~L Arnold, and Tal Arbel.
\newblock Exploring uncertainty measures in deep networks for multiple
  sclerosis lesion detection and segmentation.
\newblock {\em Medical image analysis}, 59:101557, 2020.

\bibitem{nekrasov2021mix3d}
Alexey Nekrasov, Jonas Schult, Or Litany, Bastian Leibe, and Francis Engelmann.
\newblock Mix3d: Out-of-context data augmentation for 3d scenes.
\newblock In {\em 2021 International Conference on 3D Vision (3DV)}, pages
  116--125. IEEE, 2021.

\bibitem{qi2017pointnet}
Charles~R Qi, Hao Su, Kaichun Mo, and Leonidas~J Guibas.
\newblock Pointnet: Deep learning on point sets for 3d classification and
  segmentation.
\newblock In {\em Proceedings of the IEEE conference on computer vision and
  pattern recognition}, pages 652--660, 2017.

\bibitem{qi2017pointnet++}
Charles~R Qi, Li Yi, Hao Su, and Leonidas~J Guibas.
\newblock Pointnet++: Deep hierarchical feature learning on point sets in a
  metric space.
\newblock {\em arXiv preprint arXiv:1706.02413}, 2017.

\bibitem{quinonero2008dataset}
Joaquin Qui{\~n}onero-Candela, Masashi Sugiyama, Anton Schwaighofer, and Neil~D
  Lawrence.
\newblock {\em Dataset shift in machine learning}.
\newblock Mit Press, 2008.

\bibitem{reinhold2020validating}
Jacob~C Reinhold, Yufan He, Shizhong Han, Yunqiang Chen, Dashan Gao, Junghoon
  Lee, Jerry~L Prince, and Aaron Carass.
\newblock Validating uncertainty in medical image translation.
\newblock In {\em 2020 IEEE 17th International Symposium on Biomedical Imaging
  (ISBI)}, pages 95--98. IEEE, 2020.

\bibitem{roy2019bayesian}
Abhijit~Guha Roy, Sailesh Conjeti, Nassir Navab, Christian Wachinger,
  Alzheimer's Disease~Neuroimaging Initiative, et~al.
\newblock Bayesian quicknat: Model uncertainty in deep whole-brain segmentation
  for structure-wise quality control.
\newblock {\em NeuroImage}, 195:11--22, 2019.

\bibitem{seebock2019exploiting}
Philipp Seeb{\"o}ck, Jos{\'e}~Ignacio Orlando, Thomas Schlegl, Sebastian~M
  Waldstein, Hrvoje Bogunovi{\'c}, Sophie Klimscha, Georg Langs, and Ursula
  Schmidt-Erfurth.
\newblock Exploiting epistemic uncertainty of anatomy segmentation for anomaly
  detection in retinal oct.
\newblock {\em IEEE transactions on medical imaging}, 39(1):87--98, 2019.

\bibitem{shannon1948mathematical}
Claude~Elwood Shannon.
\newblock A mathematical theory of communication.
\newblock {\em The Bell system technical journal}, 27(3):379--423, 1948.

\bibitem{smith2018understanding}
Lewis Smith and Yarin Gal.
\newblock Understanding measures of uncertainty for adversarial example
  detection.
\newblock {\em arXiv preprint arXiv:1803.08533}, 2018.

\bibitem{tang2022contrastive}
Liyao Tang, Yibing Zhan, Zhe Chen, Baosheng Yu, and Dacheng Tao.
\newblock Contrastive boundary learning for point cloud segmentation.
\newblock In {\em Proceedings of the IEEE/CVF Conference on Computer Vision and
  Pattern Recognition}, pages 8489--8499, 2022.

\bibitem{Tchapmi2017SEGCloudSS}
Lyne~P. Tchapmi, Christopher~Bongsoo Choy, Iro Armeni, JunYoung Gwak, and
  Silvio Savarese.
\newblock Segcloud: Semantic segmentation of 3d point clouds.
\newblock {\em 2017 International Conference on 3D Vision (3DV)}, pages
  537--547, 2017.

\bibitem{thomas2019KPConv}
Hugues Thomas, Charles~R. Qi, Jean-Emmanuel Deschaud, Beatriz Marcotegui,
  Fran{\c{c}}ois Goulette, and Leonidas~J. Guibas.
\newblock Kpconv: Flexible and deformable convolution for point clouds.
\newblock {\em Proceedings of the IEEE International Conference on Computer
  Vision}, 2019.

\bibitem{wang2020weakly}
Haiyan Wang, Xuejian Rong, Liang Yang, Jinglun Feng, Jizhong Xiao, and Yingli
  Tian.
\newblock Weakly supervised semantic segmentation in 3d graph-structured point
  clouds of wild scenes.
\newblock {\em arXiv preprint arXiv:2004.12498}, 2020.

\bibitem{wang2019dynamic}
Yue Wang, Yongbin Sun, Ziwei Liu, Sanjay~E Sarma, Michael~M Bronstein, and
  Justin~M Solomon.
\newblock Dynamic graph cnn for learning on point clouds.
\newblock {\em Acm Transactions On Graphics (tog)}, 38(5):1--12, 2019.

\bibitem{wei2020multi}
Jiacheng Wei, Guosheng Lin, Kim-Hui Yap, Tzu-Yi Hung, and Lihua Xie.
\newblock Multi-path region mining for weakly supervised 3d semantic
  segmentation on point clouds.
\newblock In {\em Proceedings of the IEEE/CVF Conference on Computer Vision and
  Pattern Recognition}, pages 4384--4393, 2020.

\bibitem{xu2020weakly}
Xun Xu and Gim~Hee Lee.
\newblock Weakly supervised semantic point cloud segmentation: Towards 10x
  fewer labels.
\newblock In {\em Proceedings of the IEEE/CVF Conference on Computer Vision and
  Pattern Recognition}, pages 13706--13715, 2020.

\bibitem{yan2020pointasnl}
Xu Yan, Chaoda Zheng, Zhen Li, Sheng Wang, and Shuguang Cui.
\newblock Pointasnl: Robust point clouds processing using nonlocal neural
  networks with adaptive sampling.
\newblock In {\em Proceedings of the IEEE/CVF Conference on Computer Vision and
  Pattern Recognition}, pages 5589--5598, 2020.

\bibitem{Yang_2022_CVPR}
Cheng-Kun Yang, Ji-Jia Wu, Kai-Syun Chen, Yung-Yu Chuang, and Yen-Yu Lin.
\newblock An mil-derived transformer for weakly supervised point cloud
  segmentation.
\newblock In {\em Proceedings of the IEEE/CVF Conference on Computer Vision and
  Pattern Recognition (CVPR)}, pages 11830--11839, June 2022.

\bibitem{zhang2021weakly}
Yachao Zhang, Zonghao Li, Yuan Xie, Yanyun Qu, Cuihua Li, and Tao Mei.
\newblock Weakly supervised semantic segmentation for large-scale point cloud.
\newblock In {\em Proceedings of the AAAI Conference on Artificial
  Intelligence}, volume~35, pages 3421--3429, 2021.

\bibitem{zhang2021perturbed}
Yachao Zhang, Yanyun Qu, Yuan Xie, Zonghao Li, Shanshan Zheng, and Cuihua Li.
\newblock Perturbed self-distillation: Weakly supervised large-scale point
  cloud semantic segmentation.
\newblock In {\em Proceedings of the IEEE/CVF International Conference on
  Computer Vision}, pages 15520--15528, 2021.

\bibitem{zhao2021point}
Hengshuang Zhao, Li Jiang, Jiaya Jia, Philip~HS Torr, and Vladlen Koltun.
\newblock Point transformer.
\newblock In {\em Proceedings of the IEEE/CVF International Conference on
  Computer Vision}, pages 16259--16268, 2021.

\end{thebibliography}
